\documentclass[journal]{IEEEtran}
\usepackage{amsmath,amsfonts}
\usepackage{algorithmic}
\usepackage{algorithm}
\usepackage{array}
\usepackage{textcomp}
\usepackage{stfloats}
\usepackage{url}
\usepackage{verbatim}
\usepackage{cite}
\usepackage{graphicx}
\usepackage{subcaption}
\usepackage{float}
\usepackage{booktabs}  
\usepackage{multirow}
\newcommand{\blind}{0} 

\begin{document}

\title{Graph Neural Networks Powered by Encoder Embedding for Improved Node Learning}
  \author{Shiyu Chen, Cencheng Shen, Youngser Park, Carey E. Priebe
  \thanks{
  \IEEEcompsocthanksitem Shiyu Chen is with the Department of Applied Mathematics and Statistics (AMS), Johns Hopkins University. E-mail: schen355@jh.edu \protect
  \IEEEcompsocthanksitem Cencheng Shen is with Microsoft Research. E-mail: c.shen@microsoft.com \protect
  \IEEEcompsocthanksitem Carey E. Priebe and Youngser Park
are with the Department of Applied Mathematics and Statistics (AMS), the Center for Imaging Science (CIS), and the Mathematical Institute for Data Science (MINDS), Johns Hopkins University. E-mail: cep@jhu.edu, youngser@jhu.edu \protect
}
\thanks{This work was supported in part by funding from Microsoft Research. The authors also thank the editor and anonymous reviewers for their valuable feedback.
}}



\maketitle

\begin{abstract}

Graph neural networks (GNNs) have emerged as a powerful framework for a wide range of node-level graph learning tasks. However, their performance typically depends on random or minimally informed initial feature representations, where poor initialization can lead to slower convergence and increased training instability.
In this paper, we address this limitation by leveraging a statistically grounded one-hot graph encoder embedding (GEE) as a high-quality, structure-aware initialization for node features. Integrating GEE into standard GNNs yields the GEE-powered GNN (GG) framework. Across extensive simulations and real-world benchmarks, GG provides consistent and substantial performance gains in both unsupervised and supervised settings. For node classification, we further introduce GG-C, which concatenates the outputs of GG and GEE and outperforms competing methods, achieving roughly 10–50\% accuracy improvements across most datasets. These results demonstrate the importance of principled, structure-aware initialization for improving the efficiency, stability, and overall performance of graph neural network architecture, enabling models to better exploit graph topology from the outset.

\end{abstract}


\section{Introduction}
Graphs are ubiquitous data structures that naturally capture relationships between objects, playing a crucial role in domains such as social networks, recommendation systems, and biological networks. Motivated by the pressing need to effectively learn from such graph-structured data, graph neural networks (GNNs)~\cite{kipf2016semi, kipf2016variational, hamilton2017inductive, velickovic2018graph, schlichtkrull2018modeling} have collected significant interests in recent years and delivered promising results on link prediction, node clustering, and node classification \cite{zhang2018link, zhang2021labeling, tsitsulin2023graph}.


Despite much progress in GNN research~\cite{cai2018comprehensive, wu2020comprehensive, zhang2020deep, GraphGCNSpectral}, the initialization of node embeddings remains a critical yet underexplored aspect. Most existing GNNs rely either on raw node attributes or on simplistic initialization schemes such as degree-based features, random vectors, or identity matrices when explicit attributes are absent. These initializations provide limited structural signal. Degree features capture only local connectivity and cannot distinguish nodes with different global structural roles, while identity matrices treat each node as a distinct category and effectively force the model to learn structural relationships from scratch, often resulting in slow convergence and high sensitivity to hyperparameters. Random initialization provides no structural information at all. In contrast, structure-aware initializations such as spectral embeddings and Laplacian eigenvectors offer richer global information and are supported by strong theoretical guarantees under random graph models. However, they require costly eigendecomposition, whose computational overhead can exceed that of training the GNN itself, rendering these methods impractical for large-scale graphs.  

To that end, the one-hot graph encoder embedding (GEE)~\cite{shen2022one} provides an appealing alternative that is both theoretically grounded and computationally efficient. Under standard random graph models, and given sufficiently many vertex labels, GEE is provably consistent: it converges to the latent vertex positions up to certain transformation, mirroring the theoretical guarantees of spectral embedding but without requiring any eigendecomposition. Instead of matrix factorization, GEE operates through a simple one-hot encoding of vertex labels followed by a single matrix multiplication, or equivalently, a single linear pass over the edgelist, enabling it to scale to graphs with millions of edges in seconds.

These properties make GEE a strong candidate for initializing GNNs.  Prior work has demonstrated that GEE-generated embeddings achieve competitive or superior downstream performance relative to GNNs and spectral methods, while requiring only a fraction of the computational cost, across tasks such as node classification, clustering, multi-graph embedding, latent community detection, temporal graph analysis, and outlier detection~\cite{GEERefined, GEEPrincipalCommunity, GEETemporalDynamics, GEEFusionGraphs, GEEDistanceGraph}. These evidences highlight its potential as an effective feature initialization for enhancing GNN training stability, efficiency, and predictive performance.

While GEE performs competitively in many settings, it can sometimes fall short in capturing the intricate patterns and entangled inter-class relationships present in complex real-world networks. In contrast, neural network–based approaches excel at refining local patterns and modeling complex nonlinear structures, even when initialized with random or uninformative features. This contrast suggests a natural and effective integration: combining GEE’s globally informed initialization with the GNN’s local and nonlinear refinement may produce richer embeddings and ultimately yield superior downstream performance.

In this paper, we propose and evaluate the GEE-powered GNN (GG), which incorporates GEE-generated embedding as the initial node features within the standard Graph Convolutional Network (GCN) architecture~\cite{kipf2016semi, kipf2016variational}, applied to both clustering and classification tasks. In the node clustering task, GG demonstrates not only faster convergence but also consistently superior performance compared to the baseline GNN with randomized initialization, ranking first across all 16 real-world datasets evaluated. For node classification, we further develop a concatenated variant, GG-C, which demonstrates excellent performance across a wide range of training set sizes, from as low as $5\%$ (essentially a semi-supervised setting) to $50\%$ of the nodes used for training.

The remainder of the paper is organized as follows: Section~\ref{sec: Clustering Method} introduces our implementations of the GG method based on GNN and GEE. Section~\ref{sec: motivation} provides an intuitive theoretical motivation for GG using embedding visualizations and convergence rate analyses. Section~\ref{sec: Clustering Evaluation} presents clustering performance results on both simulated and real-world datasets. Section~\ref{sec: classification} describes the classification adaptation of GG-C, which differs from the clustering setting due to label availability and changes in the learning objective. Section~\ref{sec: Classification Evaluation} reports classification results under various training percentages on both synthetic and real data. Additional experiments are provided in the supplementary material. The source code of our method is available on Github\footnote{https://github.com/chenshy202/GG}.

\section{Clustering Method} \label{sec: Clustering Method}

\subsection{GNN} \label{sec: Clustering GNN}
The architecture considered in this paper utilizes the graph convolution network with residual skip connections \cite{li2019deepgcns} and employs the loss of deep modularity networks (DMoN)\cite{tsitsulin2023graph} to enhance community structure preservation. 

\subsubsection{GCN}
The layer-wise propagation rule of GCN is defined as follows:
\begin{align*}
    Z^{(l+1)} = \sigma \left( \tilde{D}^{-\frac{1}{2}} \tilde{A} \tilde{D}^{-\frac{1}{2}} Z^{(l)} W^{(l)} \right).
\end{align*}
Here, $\tilde{A} = A + I_n$ is the adjacency matrix with added self-connections, where $A$ is the original adjacency matrix, $I_n \in {R}^{n \times n}$ is the identity matrix and $n$ is the number of nodes. The degree matrix $\tilde{D}$ of $\tilde{A}$ is defined as $\tilde{D}_{ii} = \sum_j \tilde{A}_{ij}$. $W^{(l)}$ represents the trainable weight matrix for the $l$-th layer, and $\sigma(\cdot)$ denotes the activation function, which in this case is the ReLU function, defined as $\sigma(x) = \max(0, x)$. Finally, $Z^{(l)} \in {R}^{n \times d}$ is the activation matrix in the $l$-th layer, where $Z^{(0)}$ can be either randomized inital embeddings or specific node features. Here, $d$ denotes the embedding dimension, which is set to match the number of clusters $K$.

\subsubsection{Skip connections}

Skip connections add the initial features to the outputs of each GCN layer to obtain the final representation $\hat{Z}$, which can be formulated as:
\begin{align*}
    \hat{Z} = \sum_{l=0}^{L} Z^{(l)},
\end{align*}
where $L$ denotes the number of layers, by default $2$. Skip connections are known to provide significant benefits in neural network architectures: they preserve features from earlier layers, mitigate vanishing gradients, and help reduce oversmoothing in GNNs.

\subsubsection{DMoN loss}
The loss function of DMoN combines spectral modularity maximization with a regularization term to avoid trivial solutions and is formulated as:
\begin{align*} 
C = \text{softmax}(\hat{Z}), 
\end{align*}
\begin{align*} 
\mathcal{L}_{\text{DMoN}}(C; A) = -\frac{1}{2m} \text{Tr}(C^\top B C) + \frac{\sqrt{K}}{n} \left| \left| \sum_{i} C_i^\top \right| \right|_F - 1 ,
\end{align*}
where $C \in R^{n \times K}$ is the soft community assignment matrix, obtained by applying the softmax function to the GNN output $\hat{Z}$. The adjacency matrix of the graph is denoted as $A$, and $m = \frac{1}{2} \sum_{i,j} A_{ij}$ is the total number of edges. The modularity matrix $B$ is defined as: $B = A - \frac{v v^\top}{2m}$, where $v$ represents the degree vector of the graph, computed as: $v = A \mathbf{1}_n$, with $\mathbf{1}_n \in R^n$ being a column vector of ones. 

The matrix trace $\text{Tr}(C^\top B C)$ corresponds to the modularity maximization component, which encourages dense within-class connections while reducing between-class edges. The regularization term $\left| \left| \sum_i C_i^\top \right| \right|_F$ computes the Frobenius norm of the aggregated community assignments, thereby preventing trivial solutions by mitigating the collapse of clusters.

\subsubsection{Prediction}
The soft community assignment matrix $C$, derived from the GNN output via a softmax function, serves as a probability matrix. Each node is assigned to the community with the highest probability:
\begin{align*}
\hat{Y} = \text{argmax}(C, \text{dim}=1).
\end{align*}
To formalize the entire unsupervised GNN pipeline, we formulate it as a single function $F^u_{GNN}$ that maps the initial features $Z^{(0)}$ and the graph structure $A$ to the final label predictions $\hat{Y}$:
\begin{align*}
    \hat{Y} = F^u_{GNN}(Z^{(0)}, A).
\end{align*}
In the case of the vanilla GNN, we use a randomized initial embedding $X_0$ as the input of $F^u_{GNN}$: 
\begin{align*}
\hat{Y}_{GNN} = F^u_{GNN}(X_0, A).
\end{align*}
In our implementation, $X_0 \in R^{n \times k}$ is initialized using the Xavier uniform initialization, where each embedding element is sampled from $\mathcal{U}\left( -\sqrt{\frac{6}{k + n}}, \sqrt{\frac{6}{k + n}} \right)$.

\subsection{GEE} \label{sec: Clustering GEE}
GEE is an efficient and flexible graph embedding method that can be applied with partial ground-truth labels, or labels generated by other methods, or without labels. The core step of supervised GEE is the multiplication of adjacency matrix $A$ with a column-normalized one-hot encoding matrix $W$:
\begin{align}
  Z = AW , \quad  W_{ik} =  \frac{\mathbb{I}\{\,Y_i = k\,\}}{n_k},
  \label{eq:sup_GEE}
\end{align}
where $\mathbb{I}(\cdot)$ is the indicator function and $n_k = \sum_{j=1}^{n}\mathbb{I}\{Y_j = k\}$ is the size of community $k \in [1, K]$. In practice, graph data are typically stored in edgelist format, and this step can be efficiently performed with a single linear pass over the edgelist~\cite{shen2022one, GEETemporalDynamics}, avoiding the need to construct an adjacency matrix.

Here, when node $i$ belongs to community $k$, the entry $W_{ik}$ equals $1/n_k$; Otherwise, it is $0$. Consequently, each column of $W$ sums up to $1$ and can be interpreted as a probability distribution over the nodes in that community. Eventually, the $i$-th row $Z_{i \cdot}$ summarizes how strongly node $i$ is connected to all communities, with the $k$-th column $Z_{\cdot k}$ indicating connectivity to community $k$. An interesting property of GEE is that its embedding dimension naturally matches the number of communities $K$, eliminating the need to manually select an embedding dimension.

In the clustering task, we adopt the unsupervised version of GEE. The procedure starts with a random label vector, then repeats:
\begin{enumerate}
  \item compute node embeddings with the supervised GEE, and
  \item run $k$-means on the embeddings to update the labels.
\end{enumerate}
The loop ends when the labels stop changing or when the maximum number of
iterations $M$ is reached. Formally, the whole routine can be viewed as a function $F^{u}_{GEE}(A, k, M)$: 
\begin{align}
(\hat{Z}_{GEE}, \hat{Y}_{GEE}) = F^{u}_{GEE}(A, k, M).
\label{eq:unsup_GEE}
\end{align}




    




\subsection{GEE-powered GNN (GG)} \label{sec: Clustering GG}
\begin{figure}[t]
    \centering
    \includegraphics[width=0.48\textwidth]{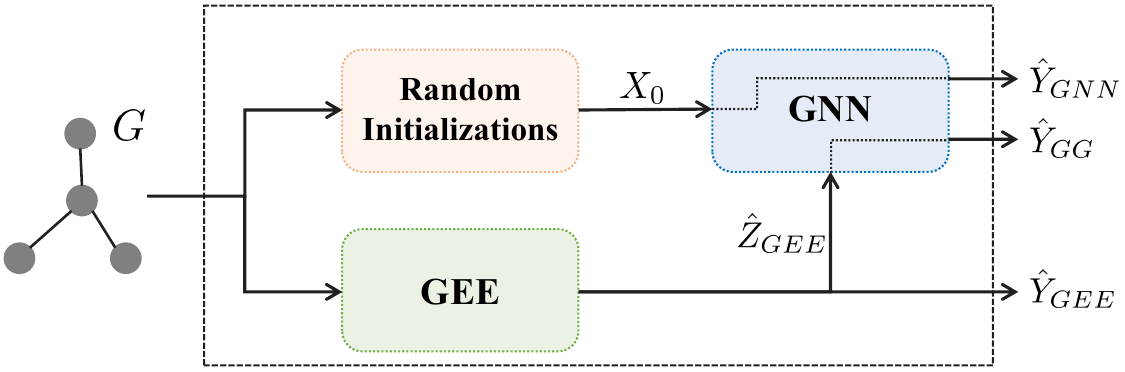}
    \caption{
        A schematic diagram of the three algorithms for the node clustering task. 
    }
    \label{fig:diagram}
\end{figure}

We generate node features $\hat{Z}_{GEE}$ from Eq.~\eqref{eq:unsup_GEE}, and utilize it as the input of the function $F^u_{GNN}$ to obtain the predicted label vector $\hat{Y}_{GG}$:
\begin{align*}
\hat{Y}_{GG} = F^u_{GNN}(\hat{Z}_{GEE}, A).
\end{align*}
Figure~\ref{fig:diagram} provides a schematic overview of the three algorithms, GNN, GEE, and GG, summarizing their workflows for the clustering task.

\begin{figure*}[t]
    \centering
    {%
        \includegraphics[width=0.32\textwidth]{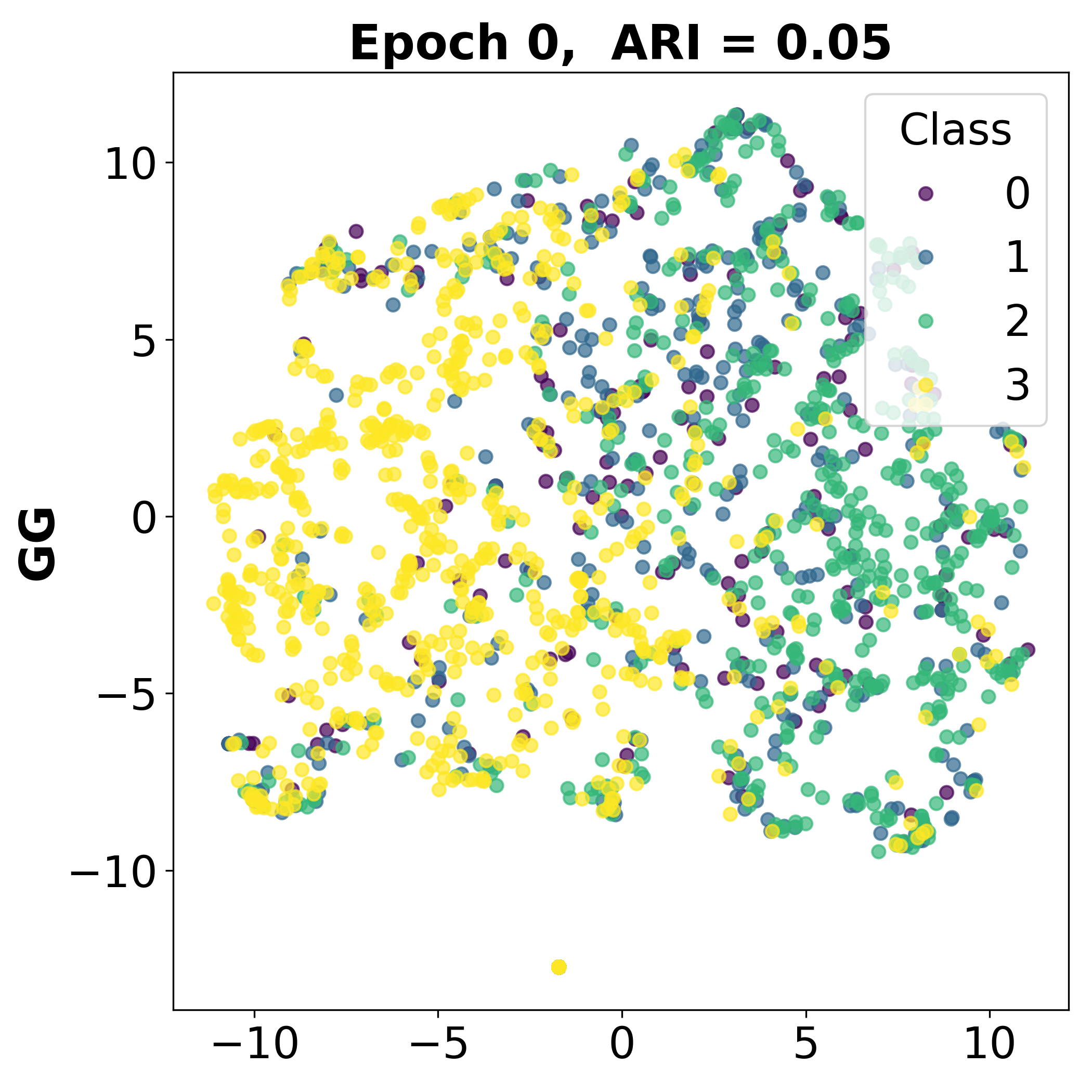}
        \label{fig:sub1}
    }
    {%
        \includegraphics[width=0.32\textwidth]{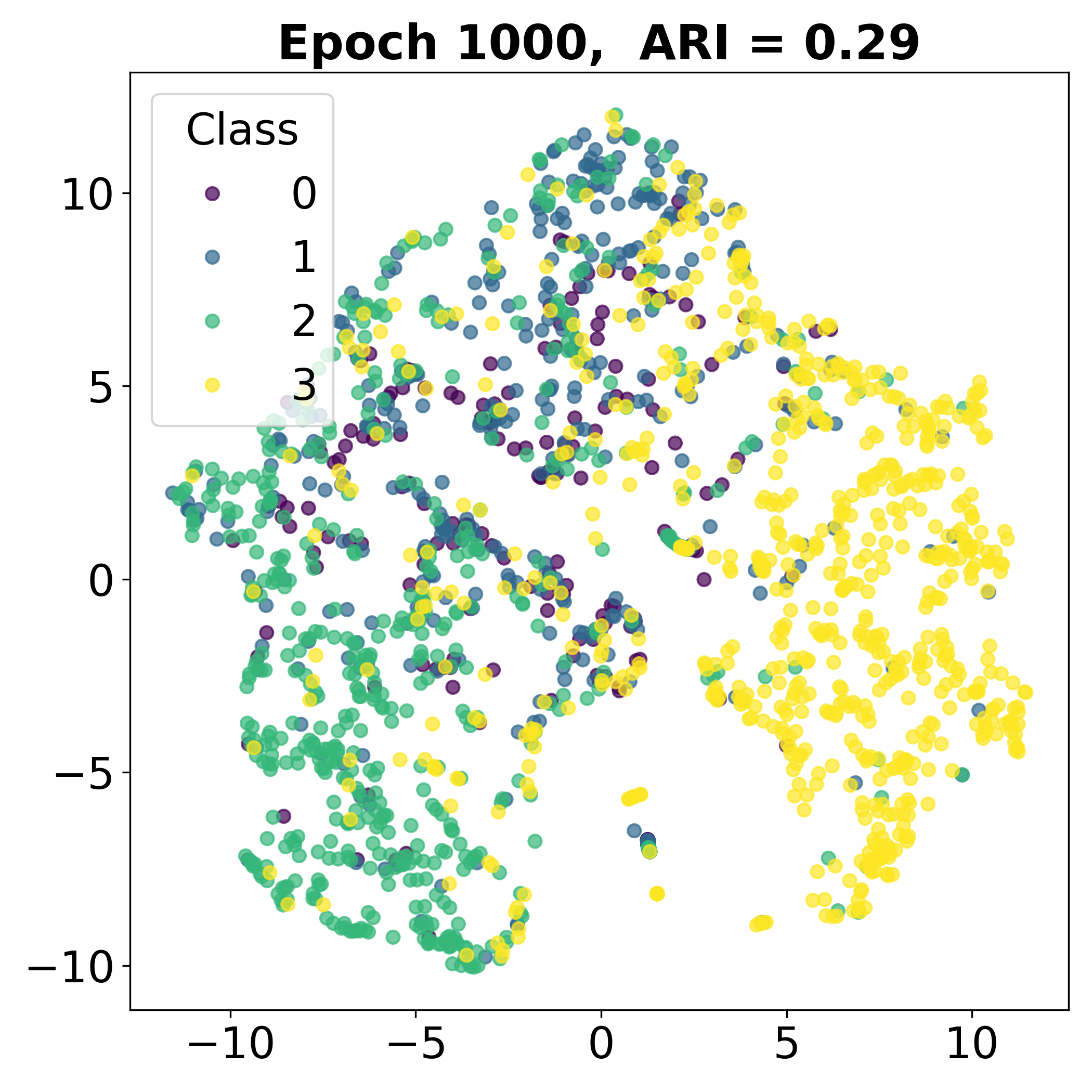}
        \label{fig:sub2}
    }
    {%
        \includegraphics[width=0.32\textwidth]{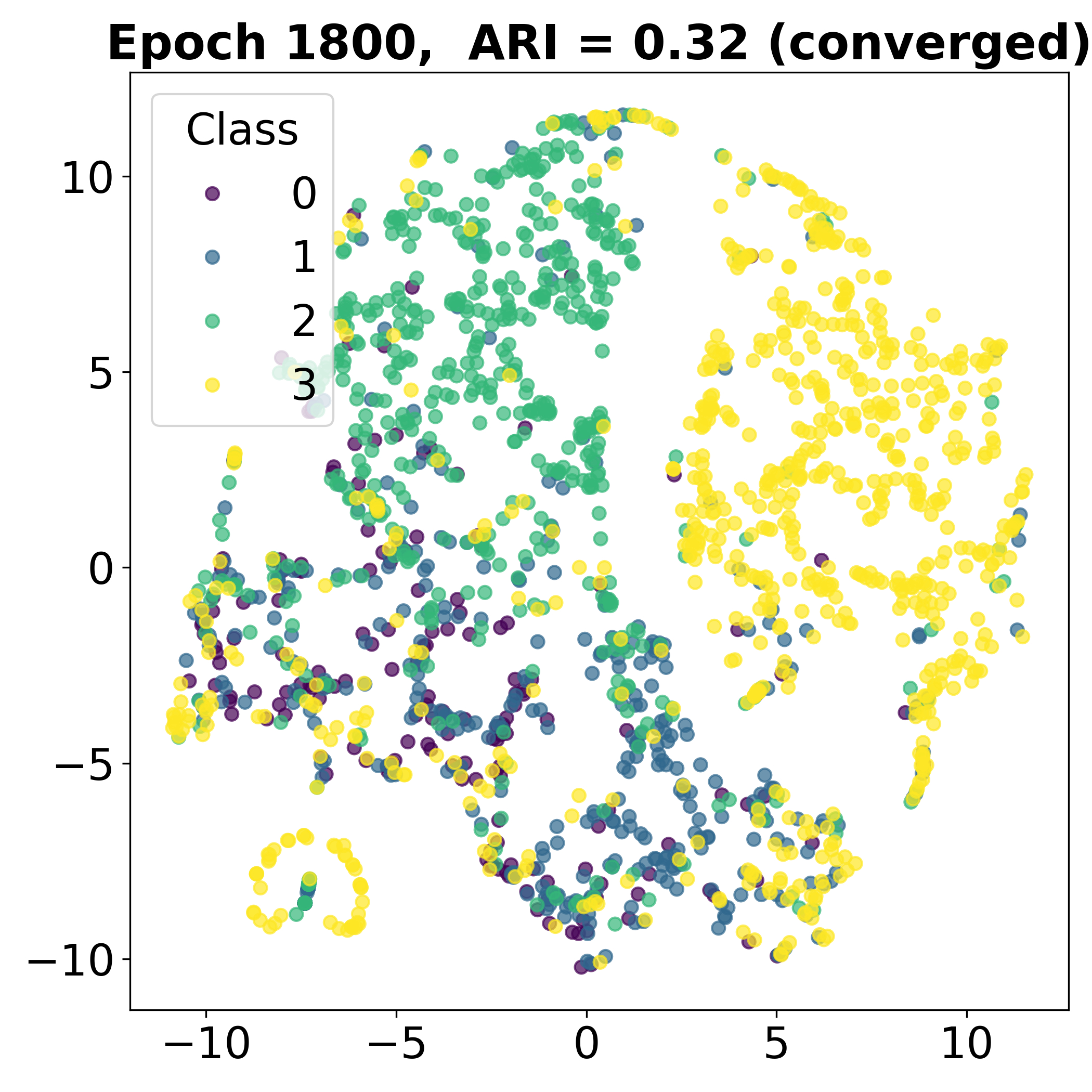}
        \label{fig:sub3}
    }
    
    {%
        \includegraphics[width=0.32\textwidth]{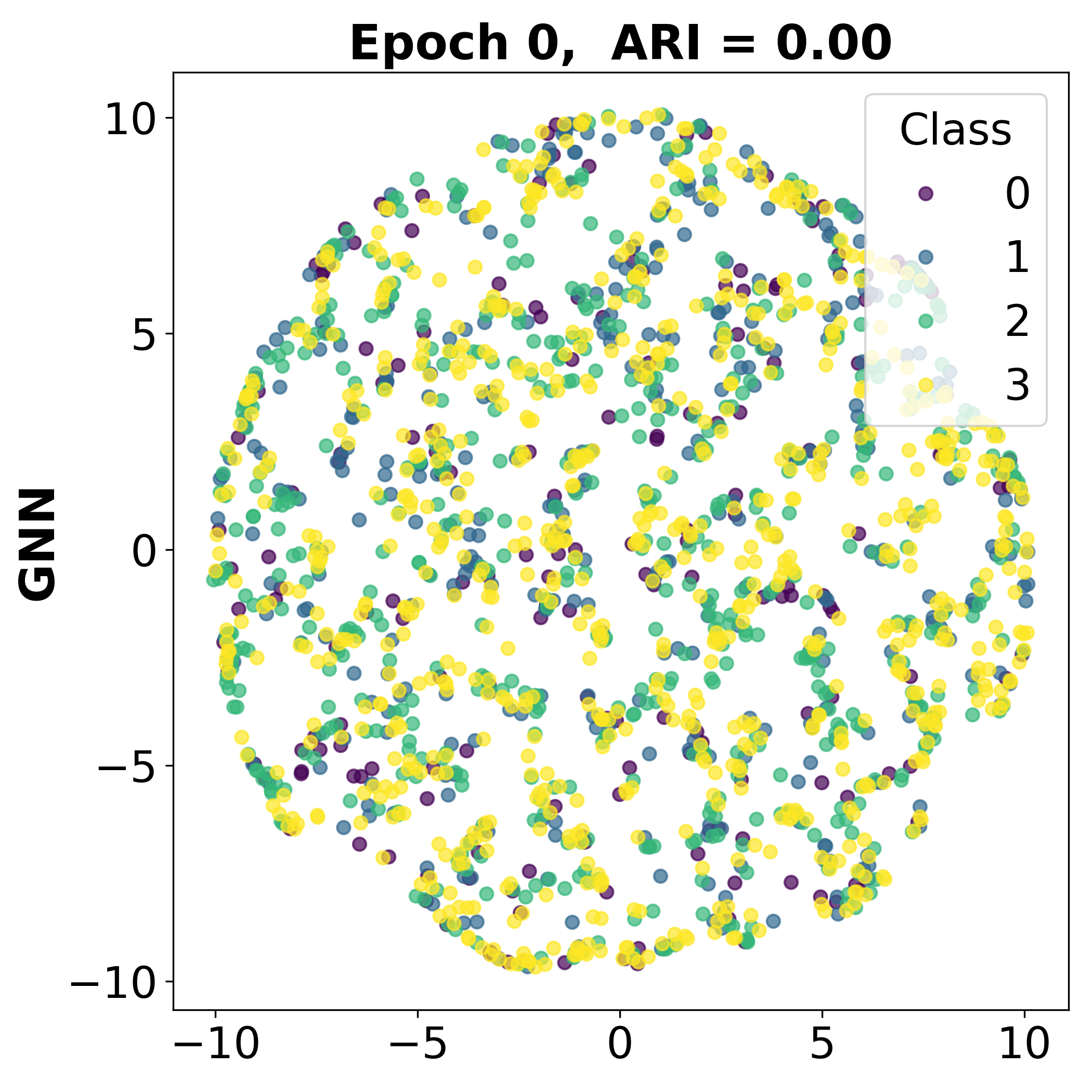}
        \label{fig:sub4}
    }
    {%
        \includegraphics[width=0.32\textwidth]{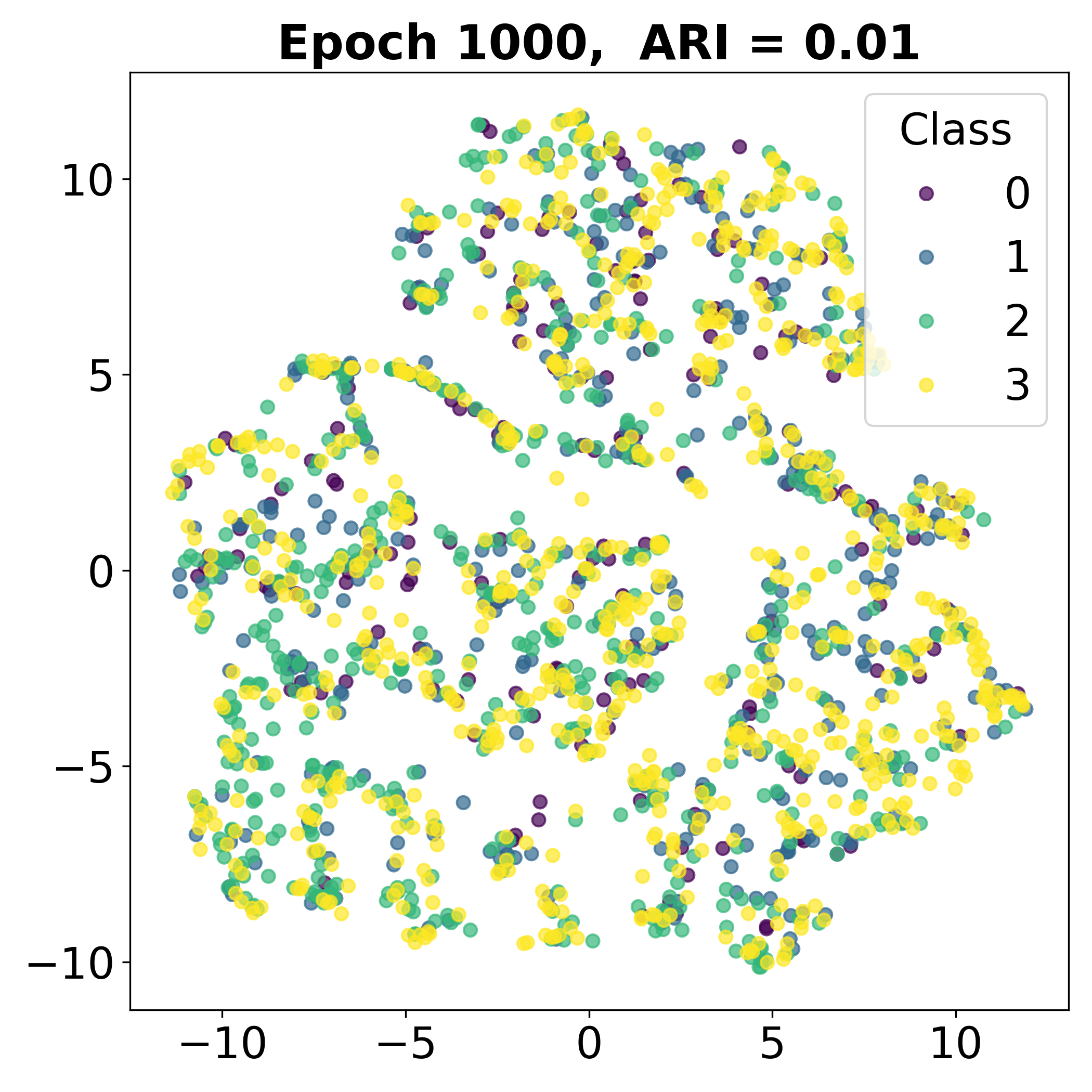}
        \label{fig:sub5}
    }
    {%
        \includegraphics[width=0.32\textwidth]{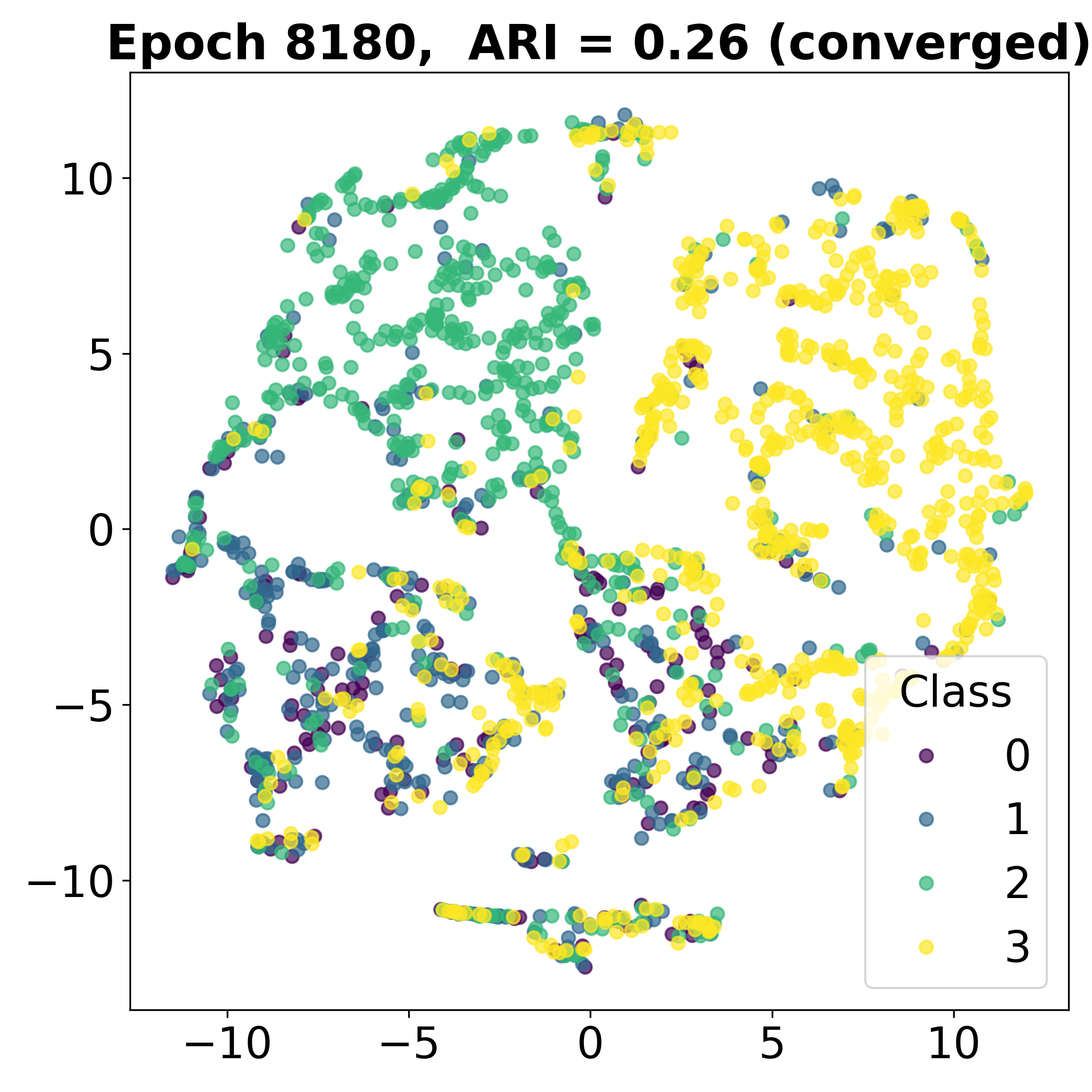}
        \label{fig:sub6}
    }
    
    \caption{
    Visualization of Training Dynamics and Clustering Quality. This figure contrasts the training process of our proposed GG (top row) and a vanilla GNN (bottom row) on a Degree-Corrected Stochastic Block Model (DC-SBM) graph (see Subsection~\ref{sec: Clustering Simulation} for detailed settings). It demonstrates that GG's warm-start leads to a better initial state, faster convergence, and a superior final clustering quality. GG begins with a structured embedding (ARI = 0.05) and converges in only 1,800 epochs to a high-quality partition (ARI = 0.32). In contrast, the GNN starts from a random state (ARI $\approx$ 0.00) and requires a much longer training of 8,180 epochs to achieve an inferior result (ARI = 0.26).
    }
    \label{fig:visualization}
\end{figure*}

\begin{figure*}[t] 
    \centering
    {
        \includegraphics[width=0.48\textwidth]{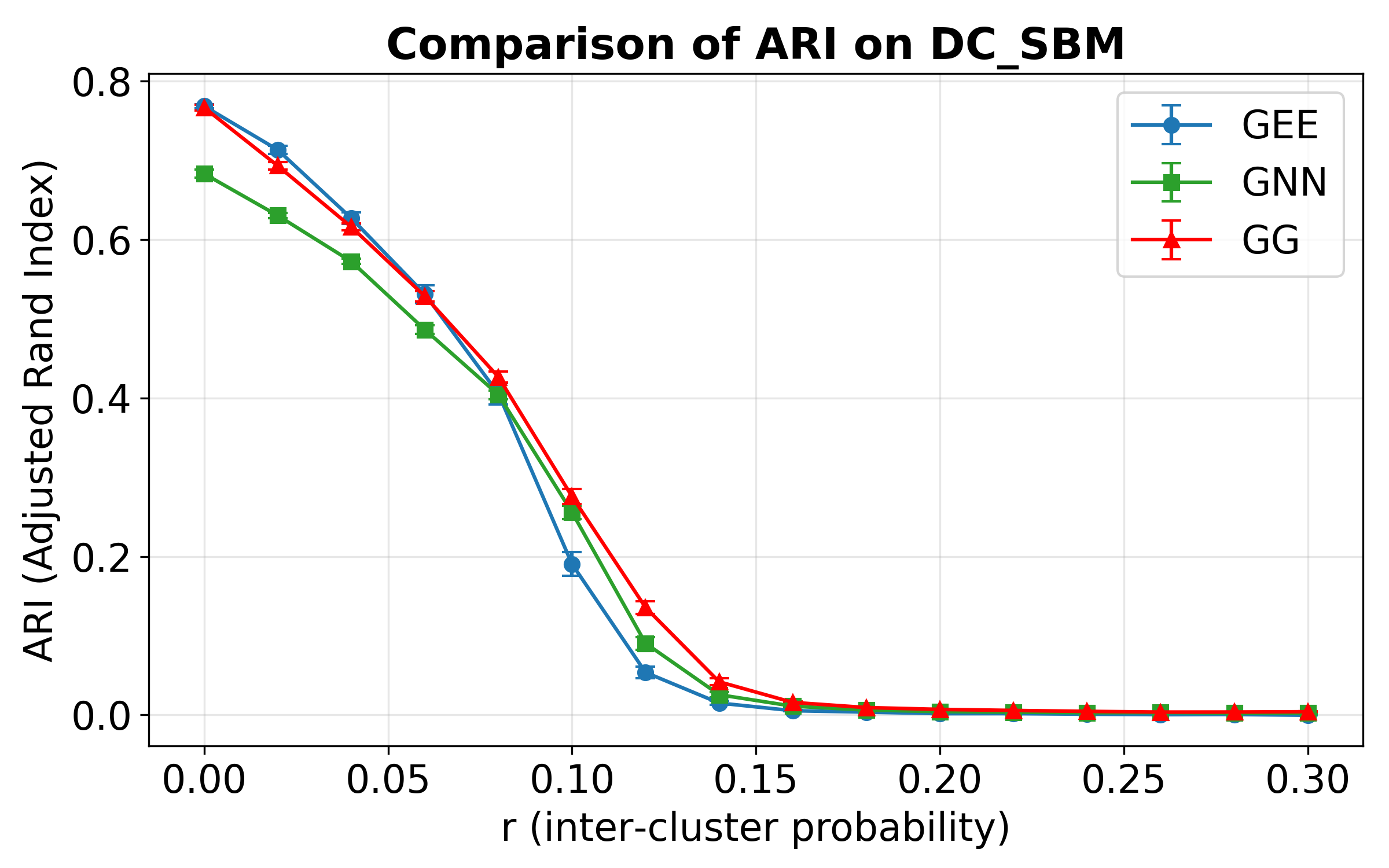}
    }
    \hfill
    {
        \includegraphics[width=0.48\textwidth]{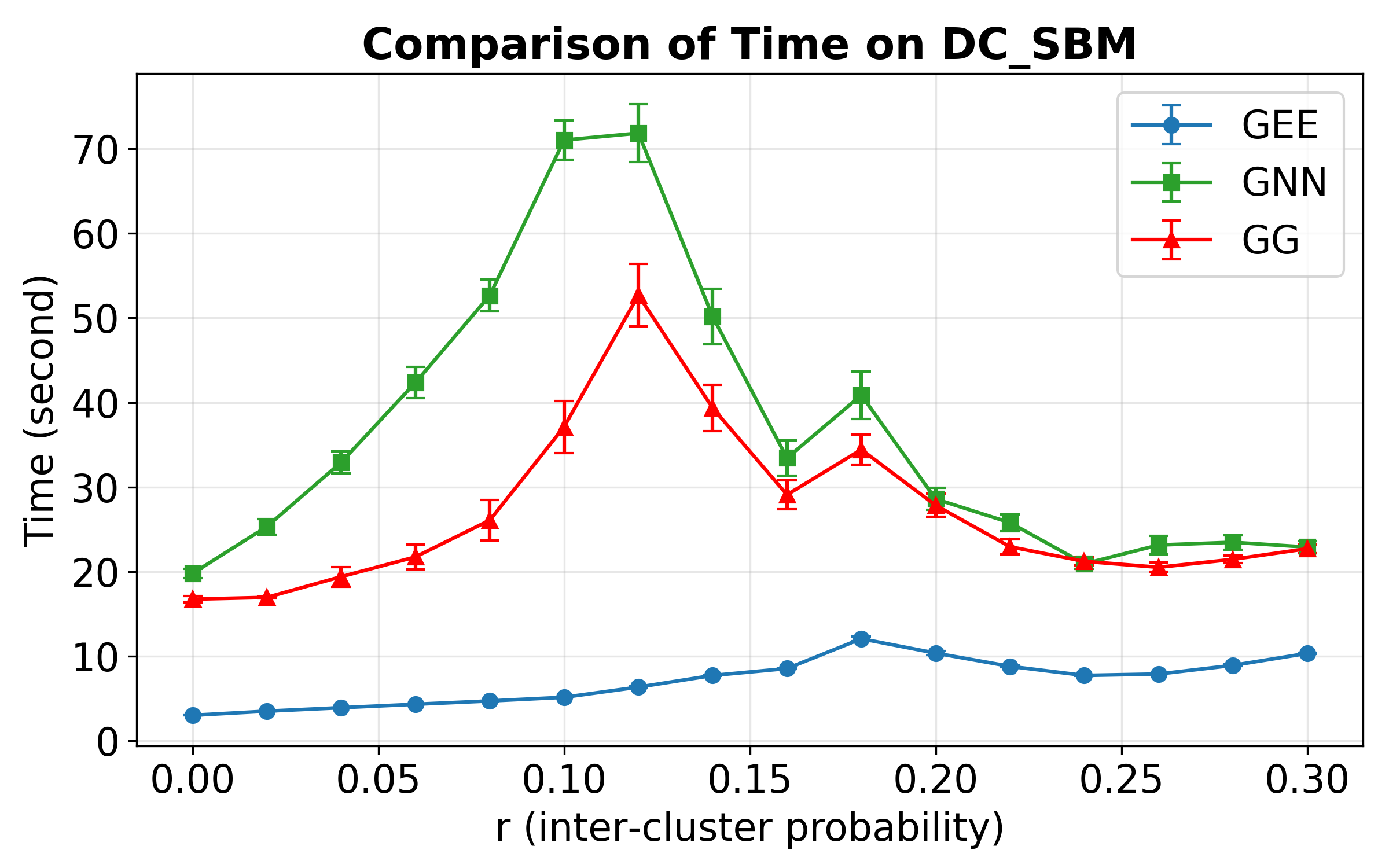}
    }
    \caption{Clustering performance comparison of GEE, GNN and GG on DC-SBM graphs. This figure shows that our proposed GG offers the best trade-off between clustering accuracy (ARI, left) and speed (Time, right). While GEE is fastest, GG matches or surpasses its accuracy, especially on more challenging graphs ($r > 0.08$). Compared to the vanilla GNN, GG is both more accurate and substantially faster across all conditions.}
    \label{fig:clustering-DC_SBM}
\end{figure*}

\section{Theoretical Motivation}
\label{sec: motivation}

Identifying meaningful community structure in graphs is a fundamental problem in network analysis~\cite{GirvanNewman2002, newman2004detecting, fortunato2010community, KarrerNewman2011}. The goal is to uncover groups of vertices that exhibit stronger intra-group connectivity than inter-group connectivity. Over the years, a wide range of approaches have been developed, including modularity-based methods~\cite{Louvain2008, Leiden2019}, spectral embedding~\cite{RoheEtAl2011, SussmanEtAl2012}, likelihood-based approaches~\cite{Gao2018, Abbe2018}, and random-walk–based methods~\cite{perozzi2014deepwalk, grover2016node2vec}, among others.

Graph neural networks differ fundamentally from the above approaches in that they optimize an objective function via gradient descent, iteratively refining an initial set of node features~\cite{kipf2016semi, kipf2016variational}. Intuitively, providing a stronger initialization can reduce the number of iterations required to reach a good solution, mitigate the risk of getting trapped in poor local minima, and potentially lead to improved downstream performance.


Recent work across diverse domains further supports the value of structurally-informed node representations. For instance, deep embedding methods such as SDNE~\cite{wang2016structural}, asymmetric transitivity preserving embeddings for directed graphs~\cite{ou2016asymmetric}, relation-symmetry based contrastive learning for knowledge graphs~\cite{liang2023knowledge}, relational multi-view clustering~\cite{liang2025concrete}, and robustness-oriented GNNs that downweight structurally uncertain neighbors~\cite{zhu2019robust} all demonstrate that explicitly encoding local, global, or relational structure can substantially improve embedding quality and downstream task performance.

This motivates our core proposal: a GEE-powered GNN, which remains a standard GNN architecture but is warm-started with node features generated by the statistically consistent GEE. By placing the initial embedding in a more informative and well-structured region of the feature space, GEE initialization potentially enables subsequent GNN training to refine representations more quickly and effectively than a vanilla GNN that relies on random or uninformative initializations~\cite{kipf2016semi}. 

To empirically validate this hypothesis, we visualize the evolution of node embeddings for both GG and the vanilla GNN during training on a DC-SBM graph. As illustrated in Figure~\ref{fig:visualization}, the impact of our warm-starting strategy is immediately apparent. GG (top row) begins at epoch 0 with some discernible community structure (ARI = 0.05), visually confirming that it starts in a "promising region" of the embedding space. Then it rapidly refines this structure, converging in only 1,800 epochs to a high-quality partition with an ARI of 0.32. In comparison, the randomly initialized GNN (bottom row) starts from a completely unstructured state (ARI = 0.00). Its learning trajectory is substantially slower, showing negligible improvement even by Epoch 1000 (ARI = 0.01). Ultimately, it requires a much longer training process (8,180 epochs) to reach a slightly inferior solution (ARI = 0.26). This side-by-side comparison offers an empirical justification for our approach. 

It is worth noting that, in principle, one could use any node embedding as an initialization for GNNs, including spectral embeddings, node2vec, or other representation-learning methods, since they all provide node features with certain theoretical guarantees. Intuitively, these warm starts should offer benefits similar to GEE; and theoretically, using any statistically consistent feature initialization ensures that the subsequent neural network refinement remains consistent as well under a proper random graph model.

However, the main limitation is computational cost: many of these embeddings require expensive matrix factorizations or optimization procedures already. In practice, it is often more efficient to simply train a GNN multiple times with random initializations and select the best outcome, than to compute a costly embedding solely for warm-starting. In contrast, GEE is both theoretically grounded and computationally lightweight, making it a practical initialization method for large-scale graphs.

\section{Clustering Evaluation} \label{sec: Clustering Evaluation}
To systematically evaluate the performance of the proposed GG method, we conduct experiments on both synthetic and real-world datasets. The evaluation considers two parts: clustering accuracy, where we compare GG against GEE and the vanilla GNN; and running time, focusing on the convergence time difference between GG and the vanilla GNN.

For each graph, whether real-world or synthetic, we generate three distinct sets of label predictions: $\hat{Y}_{GEE}$ from GEE, $\hat{Y}_{GNN}$ from a vanilla GNN, and $\hat{Y}_{GG}$ from the proposed GG. We then assess the quality of these predictions against the ground-truth labels using the Adjusted Rand Index (ARI). The ARI measures the similarity between two clusterings, where a score of $1$ indicates a perfect match, and values near $0$ suggest random partitioning.

\begin{table}[htbp]
    \centering
    \caption{Summary of datasets used in experiments}
    \label{statistic}
    {\fontsize{8.5}{11}\selectfont
    \renewcommand{\arraystretch}{1.25} 
    \begin{tabular}{lcccc}
        \toprule
        \textbf{Dataset} & \textbf{Vertices} & \textbf{Edges} & \textbf{Classes}& \textbf{Community sizes}\\
        \midrule
        Cora & 2708 & 5429 & 7 & 818, 426, 418, 351\\
        Citeseer & 3327 & 4732 & 6 & 701, 668, 596, 590\\
        ACM & 3025 & 13128 & 3 & 1061, 999, 965\\
        Chameleon & 2277 & 18050 & 5 & 521, 460, 456, 453\\
        DBLP & 4057 & 3528 & 4 & 1197, 1109, 1006, 745\\
        EAT & 399 & 203 & 4 & 102, 99,  99,  99\\
        UAT & 1190 & 13599 & 4 & 299, 297, 297, 297\\
        BAT & 131 & 1074 & 4 & 35, 32, 32, 32\\
        KarateClub & 34 & 156 & 4 & 13, 12, 5, 4\\
        Email & 1005 & 25571 & 42 & 109, 92, 65, 61\\
        Gene & 1103 & 1672 & 2 & 613, 490\\
        Industry & 219 & 630 & 3 & 133, 44, 42\\
        LastFM & 7624 & 27806 & 17 & 1303, 1098, 655, 515\\
        PolBlogs & 1490 & 33433 & 2 & 636, 588\\
        Wiki & 2405 & 16523 & 17 & 406, 361, 269, 228\\
        TerroristRel & 881 & 8592 & 3 & 633, 181, 67\\
        \bottomrule
    \end{tabular}
    }
\end{table}

\subsection{Simulation} \label{sec: Clustering Simulation}

For simulations, we consider the stochastic block model (SBM)~\cite{holland1983stochastic} and the degree-corrected SBM (DC-SBM)~\cite{zhao2012consistency}. Here we present the results on DC-SBM, as it better captures the degree heterogeneity in real-world networks. Results for SBM are available in Appendix.

In DC-SBM, the generative process for an edge between two nodes depends on three components: their community assignments, a shared block probability matrix, and individual degree parameters. First, each node $i$ is assigned to a community, represented by a label $Y_i \in \{1, \dots, K\}$. The connectivity between these communities is governed by the block probability matrix $B \in [0,1]^{K \times K}$, where entry $B(k,l)$ specifies the probability of a link between a node in community $k$ and a node in community $l$. The key distinction from the standard SBM is the introduction of a degree parameter $\theta_i$ for each node to account for degree heterogeneity.

Combining these elements, the adjacency matrix $ A \in \{0,1\}^{n \times n} $ is generated as follows:
\begin{align*}
   A_{ij} \sim \text{Bernoulli}\left(\theta_i \theta_j B(Y_i,Y_j)\right)
\end{align*}
where $ A(i,i) = 0 $ and $ A(j,i) = A(i,j) $ to ensure the graph is simple and undirected.

For the experimental setup, the graphs are configured with $n=2000$ nodes and $K=4$ communities. To evaluate model performance under varying degrees of community separability, we fix the intra-community edge probability at $0.3$ while systematically varying the inter-community probability $r$ from $0$ to $0.3$ in increments of $0.02$. To introduce degree heterogeneity, the degree correction parameters $\theta_i$ are drawn from a Beta distribution, $\beta(1,4)$. Furthermore, we impose a structural imbalance by setting the community sizes to a ratio of 1:2:3:4, a choice intended to better reflect the skewed community size distributions often found in real-world networks.

To ensure the statistical reliability of our findings across all simulation experiments (including SBM in Appendix), we conduct 50 independent runs for each graph generated at every value of $r$ for all simulations. We then report the mean and standard error of the performance and running time, which are visualized as error bars in the result plots. The visualizations of node embeddings presented in Figure~\ref{fig:visualization} are taken from a representative run of this experimental setup, specifically using a graph generated with $r=0.1$.

\begin{table*}[t]
  \centering
  \caption{Clustering accuracy (ARI) and running time on real-world datasets. The best result is highlighted in \textbf{bold}. Running time for GEE is excluded from comparison with GNN-based models due to its non-iterative nature.}
  \label{tab:Clustering real data}
   \begin{subtable}[t]{\textwidth}
        \centering
        \label{tab:high_ari}
        \renewcommand{\arraystretch}{1.0}
        \small
        \begin{tabular}{lcccccccc}
            \toprule
            & UAT & Industry & Wiki & Cora & LastFM & Email & KarateClub & PolBlogs \\
            \midrule
            \multicolumn{9}{c}{ARI} \\
            \midrule
            GEE & 0.090$\pm$0.001 & 0.061$\pm$0.006 & 0.116$\pm$0.002 & 0.137$\pm$0.004 & 0.341$\pm$0.004 & 0.436$\pm$0.005 & 0.449$\pm$0.014 & 0.812$\pm$0.001 \\
            GNN & 0.126$\pm$0.002 & 0.099$\pm$0.005 & 0.106$\pm$0.003 & 0.111$\pm$0.004 & 0.274$\pm$0.009 & 0.340$\pm$0.003 & 0.725$\pm$0.011 & 0.745$\pm$0.003 \\
            GG  & \textbf{0.140$\pm$0.003} & \textbf{0.126$\pm$0.005} & \textbf{0.124$\pm$0.002} & \textbf{0.166$\pm$0.004} & \textbf{0.409$\pm$0.004} & \textbf{0.537$\pm$0.002} & \textbf{0.726$\pm$0.013} & \textbf{0.821$\pm$0.003} \\
            \midrule
            \multicolumn{9}{c}{Running time (s)} \\
            \midrule
            GEE & 5.9$\pm$0.0 & 0.6$\pm$0.0 & 7.1$\pm$0.0 & 3.4$\pm$0.0 & 14.2$\pm$0.1 & 8.3$\pm$0.03 & 0.3$\pm$0.0 & 6.6$\pm$0.1 \\
            GNN & \textbf{12.0$\pm$0.2} & \textbf{2.8$\pm$0.1} & 104.6$\pm$5.1 & 80.6$\pm$2.5 & 779.6$\pm$48.1 & 29.4$\pm$0.4 & 1.8$\pm$0.0 & 17.5$\pm$1.0 \\
            GG  & 12.4$\pm$0.3 & 2.9$\pm$0.1 & \textbf{65.1$\pm$3.7} & \textbf{59.2$\pm$3.7} & \textbf{474.1$\pm$29.9} & \textbf{28.2$\pm$0.3} & \textbf{1.8$\pm$0.0} & \textbf{10.4$\pm$0.1} \\
    
            \bottomrule
        \end{tabular}
   \end{subtable}

    \vspace{1em}
    
   \begin{subtable}[t]{\textwidth}
        \centering
        \label{tab:low_ari}
        \renewcommand{\arraystretch}{1.0}
        \small
        \begin{tabular}{lcccccccc}
            \toprule
            & Gene & TerroristRel & DBLP & EAT & Citeseer & ACM & Chameleon & BAT \\
            \midrule
            \multicolumn{9}{c}{ARI} \\
            \midrule
            GEE & 0.008$\pm$0.002 & 0.062$\pm$0.008 & 0.018$\pm$0.001 & 0.017$\pm$0.001 & 0.046$\pm$0.003 & 0.033$\pm$0.003 & 0.045$\pm$0.002 & 0.043$\pm$0.003 \\
            GNN & 0.008$\pm$0.002 & 0.075$\pm$0.005 & 0.009$\pm$0.001 & 0.023$\pm$0.002 & 0.035$\pm$0.002 & 0.034$\pm$0.003 & 0.056$\pm$0.002 & 0.060$\pm$0.003 \\
            GG  & \textbf{0.011$\pm$0.002} & \textbf{0.078$\pm$0.009} & \textbf{0.022$\pm$0.001} & \textbf{0.044$\pm$0.005} & \textbf{0.056$\pm$0.003} & \textbf{0.059$\pm$0.003} & \textbf{0.060$\pm$0.002} & \textbf{0.085$\pm$0.007} \\
            \midrule
            \multicolumn{9}{c}{Running time (s)} \\
            \midrule
            GEE & 1.3$\pm$0.0 & 2.5$\pm$0.1 & 2.6$\pm$0.0 & 2.6$\pm$0.0 & 3.2$\pm$0.0 & 5.8$\pm$0.0 & 13.0$\pm$0.1 & 0.7$\pm$0.0 \\
            GNN & 8.4$\pm$0.3 & 7.9$\pm$0.2 & 85.9$\pm$5.2 & 4.6$\pm$0.1 & 93.7$\pm$3.7 & 54.0$\pm$2.9 & 44.6$\pm$2.1 & 2.1$\pm$0.0 \\
            GG  & \textbf{7.9$\pm$0.1} & \textbf{7.8$\pm$0.2} & \textbf{70.3$\pm$3.2} & \textbf{4.5$\pm$0.1} & \textbf{63.5$\pm$3.0} & \textbf{47.8$\pm$2.8} & \textbf{31.3$\pm$1.0} & \textbf{2.1$\pm$0.0} \\
            \bottomrule
        \end{tabular}
   \end{subtable}

\end{table*}

Figure~\ref{fig:clustering-DC_SBM} summarizes the performance. In terms of clustering accuracy (left panel), the results reveal the distinct roles of each component. As anticipated, GEE demonstrates high accuracy when the community structure is clear (low $r$), which degrades as the network becomes more complex ($r>0.1$). The vanilla GNN, while performing more consistently than GEE in this regime, yields suboptimal results. This is the exact scenario where GG is designed to solve, which consistently outperforms the vanilla GNN across the entire difficulty spectrum. This demonstrates that GEE's high-quality initialization allows the GNN component to avoid poor local optima and fine-tune its way to a superior solution. Regarding computational time (right panel), GEE is the fastest method, the vanilla GNN is the most costly, while GG strikes a fine balance, costing less than vanilla due to the better initialization. 

\subsection{Real Data} \label{sec: Clustering Real data Experiments}
We evaluate the performance of the proposed method on a diverse suite of 16 publicly available real-world graphs. Following the methodology of prior work~\cite{shen2022one, liu2022survey, wang2023overview}, we select datasets from public sources\footnote{Available: \url{https://github.com/cshen6/GraphEmd} and \url{https://github.com/yueliu1999/Awesome-Deep-Graph-Clustering}}. This collection includes benchmark citation networks (e.g., Cora, Citeseer), co-authorship graphs (e.g., ACM, DBLP), social and communication networks (e.g., PolBlogs, LastFM, Email), web graphs based on hyperlinks (e.g., Chameleon, Wiki), biological interaction networks (e.g., Gene) and others representing a variety of structural properties and community distribution patterns.


The key statistics of these datasets are summarized in Table~\ref{statistic}. The table also details the size of each community within the datasets. For datasets with more than four communities, the sizes of the four largest communities are listed. As shown in the table, imbalanced community sizes are a common characteristic of these networks. To ensure reliable results, all experiments are repeated for 50 independent trials, and we report the mean and standard error for both ARI and running time. The detailed results are presented in Tables~\ref{tab:Clustering real data}, with the best ARI score for each dataset highlighted in bold. 
The results clearly demonstrate the advantages of GG. Not only does it achieve the best accuracy on all real-world datasets, but it also operates with substantially less computational time than the vanilla GNN in most cases. This empirical observation aligns closely with the simulation results. 

\section{Classification Method}
\label{sec: classification}
\subsection{GNN} \label{sec: Classification GNN}
For node classification, we employ the same graph convolution network architecture except a different loss function. The layer-wise update follows as:
\begin{align*}
    Z^{(l+1)} &= \sigma \left( \tilde{D}^{-\frac{1}{2}} \tilde{A} \tilde{D}^{-\frac{1}{2}} Z^{(l)} W^{(l)} \right),\\
    \hat{Z} &= \sum_{l=0}^{L} Z^{(l)}.
\end{align*}
The model is trained by minimizing the cross-entropy loss $\mathcal{L}_{\text{CE}}$, over the set of training nodes $\mathcal{V}_{\text{train}}$. Given the output $\hat{Z}$, and ground-truth labels $Y$, the loss is defined as:
\begin{align}
    \mathcal{L}_{\text{CE}} = - \frac{1}{|\mathcal{V}_\text{train}|} \sum_{i \in \mathcal{V}_\text{train}} \log \left( \frac{\exp(\hat{Z}_{iY_i})}{\sum_{c=1}^{K} \exp(\hat{Z}_{ic})} \right).
    \label{eq:cross_entropy_compact}
\end{align}
Once trained, the model produces the final prediction $\hat{Y}$. The predicted class label for each node $i$ is then determined by applying the argmax operator: 
\begin{align*}
    \hat{Y_i} = \arg\max_{c} (\hat{Z}_{ic}), \quad c \in [0, K-1].
\end{align*}

To represent this entire end-to-end process, from the initial node representations $Z^{(0)}$ and the adjacency matrix $A$ to the final label predictions, we define a single function $F^a_{GNN}$, The output is thus expressed as:
\begin{align*}
\hat{Y} = F^a_{GNN}(Z^{(0)}, A).
\end{align*}

For the vanilla GNN baseline, we use the same random initialization scheme as in the clustering task, generating an initial embedding matrix $X_0$ as the input of $F^u_{GNN}$: 
\begin{align*}
\hat{Y}_{GNN} = F^a_{GNN}(X_0, A).
\end{align*}

\subsection{GEE} \label{sec: Classification GEE}
For GEE in the classification task, we generate node embeddings $\hat{Z}_{GEE}$ using the supervised GEE algorithm (Eq.~\eqref{eq:sup_GEE}), where the labels for test nodes $\mathcal{V}_\text{test}$ are masked (e.g., set to $-1$) and not used by the GEE algorithm. We denote the masked label vector $\tilde{Y}$ as:
\begin{align}
  \tilde{Y}_i =
  \begin{cases}
    Y_i & \text{if } i \in \mathcal{V}_{\text{train}} \\
    -1  & \text{if } i \in \mathcal{V}_{\text{test}}
  \end{cases}
  \label{eq:masked_labels}
\end{align}
Using this masked label vector, the supervised GEE embeddings are then computed as: 
\begin{align}
    \hat{Z}_{GEE} = AW, \quad W_{ik} =  \frac{\mathbb{I}\{\,\tilde{Y_i} = k\,\}}{n_k}.
\label{eq:sup_GEE2}
\end{align}
To output labels for the test nodes, we use the linear discriminant analysis (LDA) on $\hat{Z}_{GEE}$:
\begin{align}
\hat{Y}_{GEE} = \text{LDA}(\hat{Z}_{GEE}).
\label{eq:cl_GEE}
\end{align}

\begin{figure*}[t] 
    \centering
    {
        \includegraphics[width=0.45\textwidth]{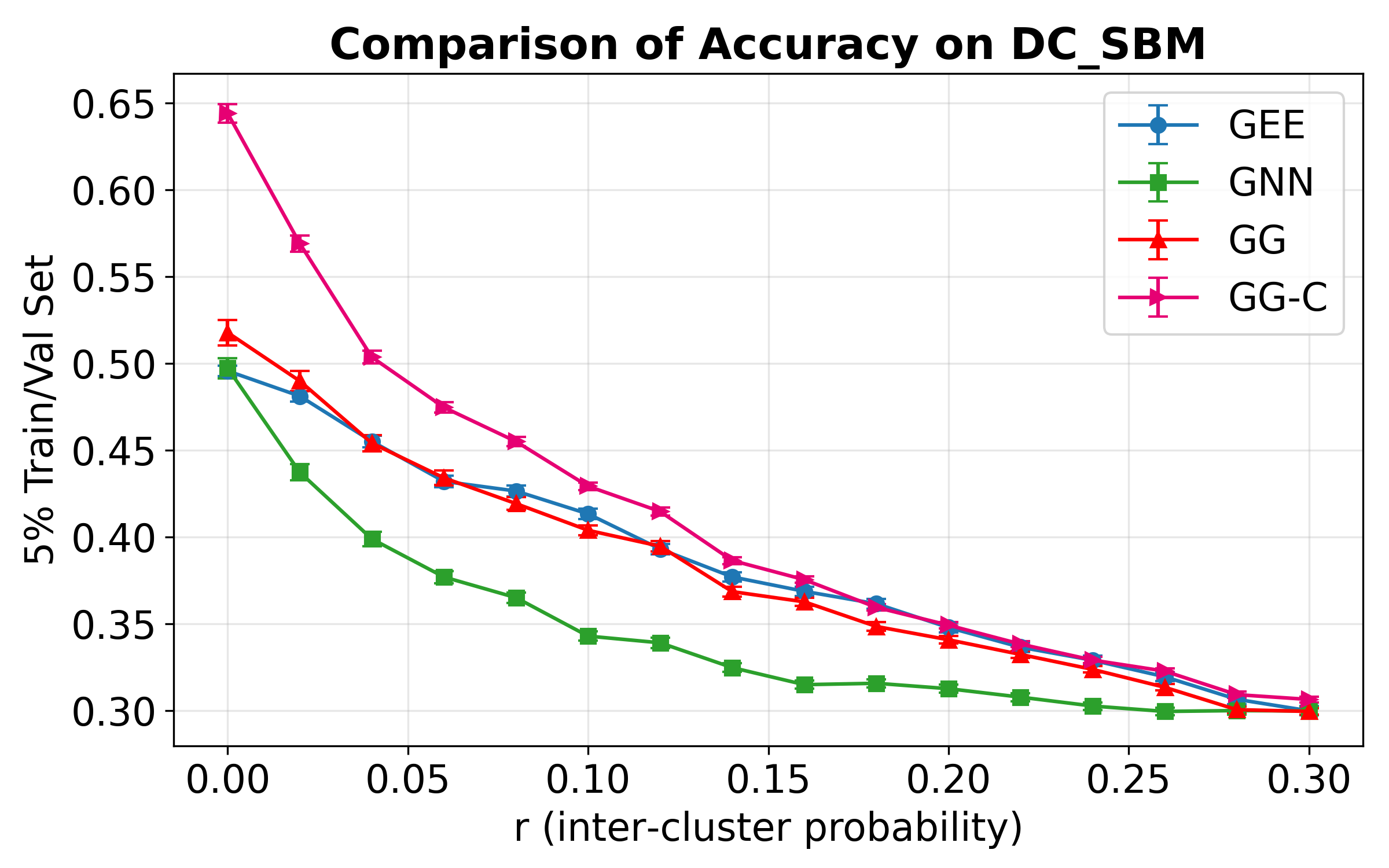}
    }
    \hfill
    {
        \includegraphics[width=0.45\textwidth]{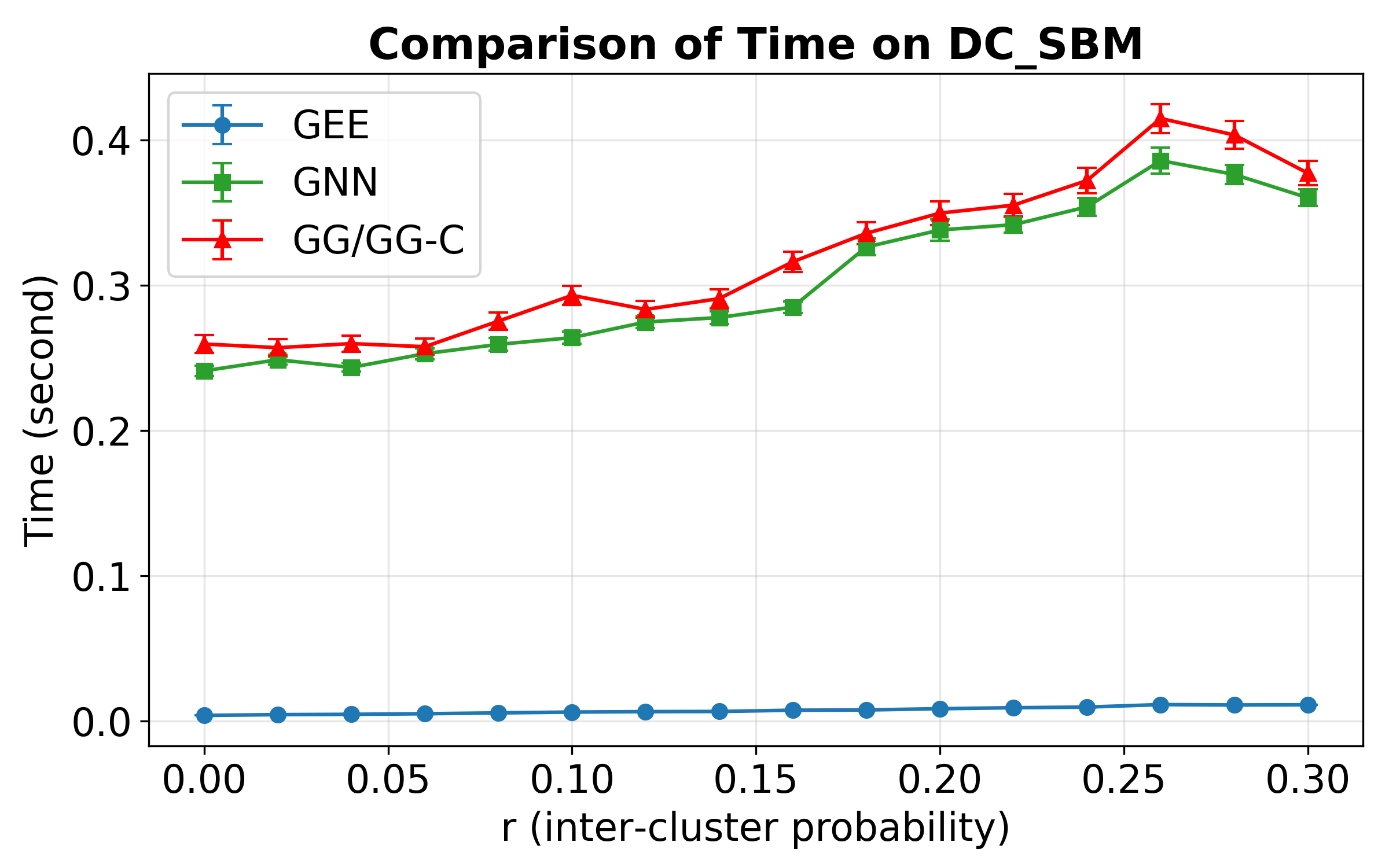}
    }
    \centering
    {
        \includegraphics[width=0.45\textwidth]{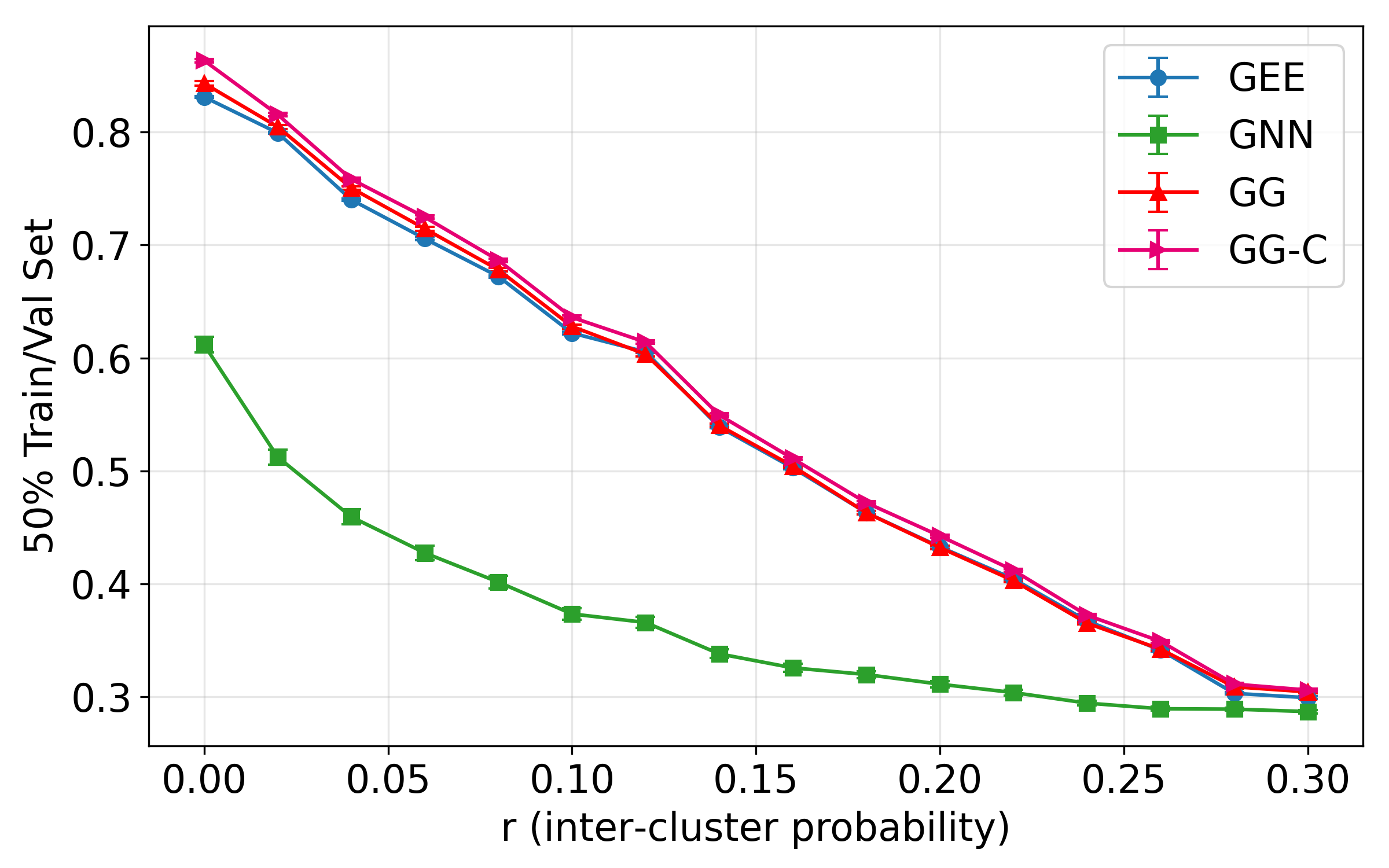}
    }
    \hfill
    {
        \includegraphics[width=0.45\textwidth]{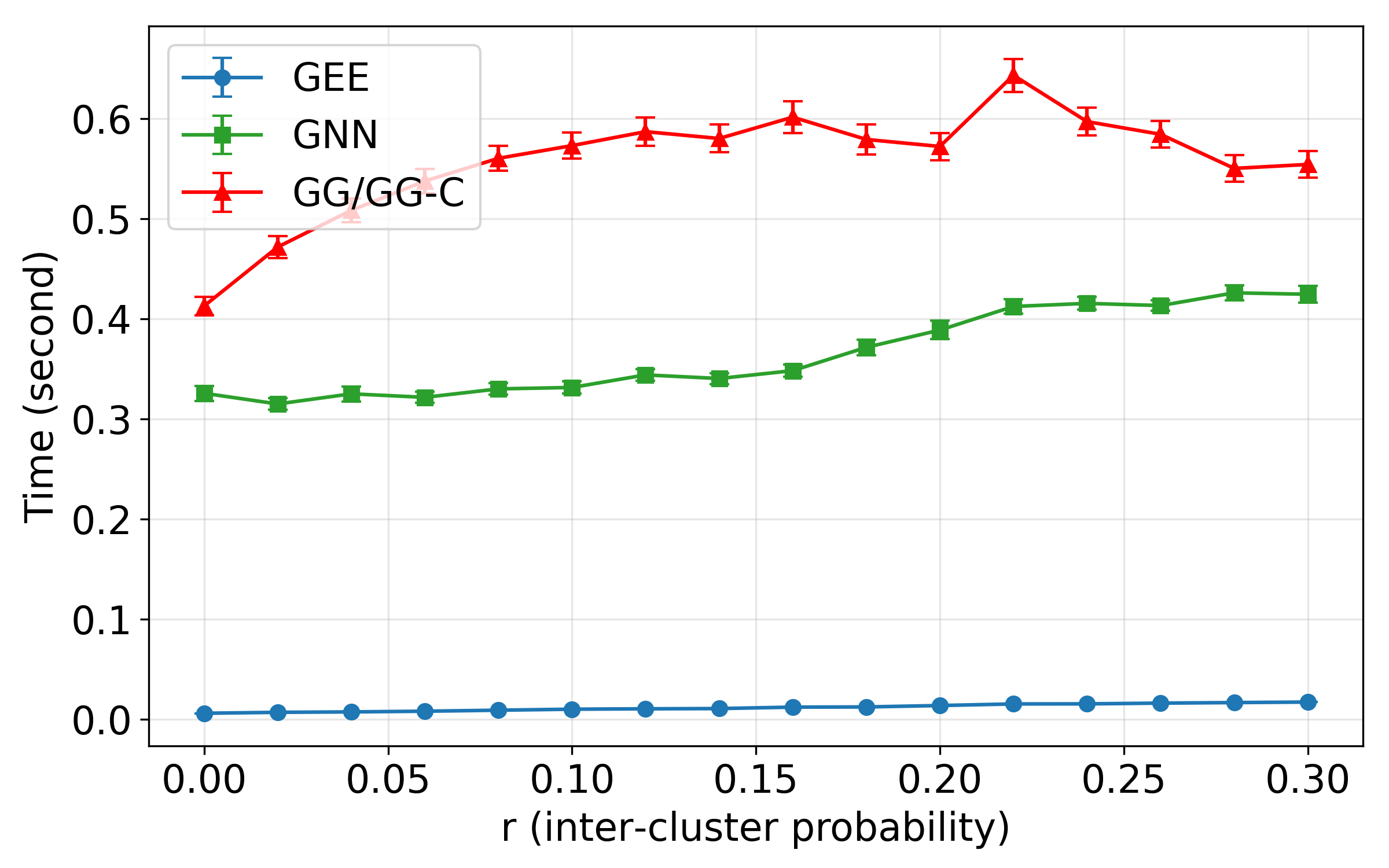}
    }
    \caption{Classification performance comparison of GEE, GNN, GG and GG-C on DC-SBM graphs with 5\% (top row) and 50\% (bottom row) train/val set. This figure demonstrates the superiority of our GG and GG-C. They overcome the poor accuracy of the faster GNN (which is prone to local optima), while our concatenated variant, GG-C, also surpasses the classic GEE, particularly in the challenging low-label (5\%) setting. }
    \label{fig:classification DC_SBM 5 and 50}
\end{figure*}

\subsection{GG} \label{sec: Classification GG}
Following the same warm-start strategy in the clustering task, the GG method for classification initializes GNN with features derived from GEE. Specifically, we utilize the supervised GEE embeddings, $\hat{Z}_{GEE}$ (from Eq.~\eqref{eq:sup_GEE2}) as input of the function $F^a_{GNN}$ to obtain the predicted label vector $\hat{Y}_{GG}'$:
\begin{align*}
\hat{Y}_{GG} = F^a_{GNN}(\hat{Z}_{GEE}, A).
\end{align*}

\subsection{GG-C} \label{sec: Classification GG-C}
Recognizing that GEE and GG may capture complementary aspects of the graph, where GEE emphasizing global structure and GG refining local patterns, we introduce a variant termed GG-C, which concatenates the respective embeddings to form an even richer representation for each node. Specifically, we concatenate the output of the supervised GEE, denoted as $\hat{Z}_{GEE}$, with the final GNN-refined embeddings from the GG model, denoted as $\hat{Z}_{GG}$. The concatenated embedding is then passed to an LDA classifier for final label prediction. Formally, the prediction can be expressed as:
\begin{align*}
\hat{Y}_{GG-C} = \text{LDA}([\hat{Z}_{GG},\hat{Z}_{GEE}]).
\end{align*}
This design is particularly advantageous for node classification. If either embedding component contains little discriminative information, the LDA classifier remains unaffected; if one or both components offer additional informative signals, whether individually or jointly, LDA is well-positioned to extract and leverage them. This idea is conceptually similar to the synergistic effect of multiple graph representations using GEE~\cite{GEEFusionGraphs}.

\begin{table*}[htbp]
  \centering
  \caption{Classification accuracy and running time on real-world datasets of 5\% (top) and 50\% (bottom) train/val data. For accuracy, the best and second-best results are highlighted in \textbf{bold} and \textit{italic}, respectively. For the GNN-based models (GNN, GG and GG-C), the fastest running time is also marked in bold. The running time of GEE is not highlighted and is excluded from direct comparison due to its non-iterative nature.}
  \label{tab:classification realdata 5 and 50}
  \begin{subtable}[t]{\textwidth}
    \centering
    \label{tab:part1}
    \begin{tabular}{lcccccccc}
      \toprule
      5\% & ACM & BAT & DBLP & EAT & UAT & Wiki & Gene & IIP \\
      \midrule
      \multicolumn{9}{c}{Accuracy}\\
      \midrule
      GEE & 0.469$\pm$0.002 & \textit{0.412$\pm$0.006} & 0.316$\pm$0.001 & \textit{0.364$\pm$0.005} & \textit{0.415$\pm$0.005} & 0.244$\pm$0.004 & \textbf{0.575$\pm$0.004} & 0.410$\pm$0.014 \\
      GNN & 0.427$\pm$0.004 & 0.294$\pm$0.005 & 0.297$\pm$0.002 & 0.321$\pm$0.005 & 0.358$\pm$0.005 & 0.224$\pm$0.003 & 0.525$\pm$0.006 & \textbf{0.515$\pm$0.008} \\
      GG  & \textit{0.492$\pm$0.005} & 0.382$\pm$0.007 & \textit{0.329$\pm$0.002} & 0.351$\pm$0.005 & 0.404$\pm$0.006 & \textbf{0.326$\pm$0.003} & 0.531$\pm$0.005 & \textit{0.510$\pm$0.010} \\
      GG-C & \textbf{0.534$\pm$0.003} & \textbf{0.425$\pm$0.008} & \textbf{0.336$\pm$0.003} & \textbf{0.370$\pm$0.005} & \textbf{0.435$\pm$0.004} & \textit{0.304$\pm$0.003} & \textit{0.549$\pm$0.003} & 0.444$\pm$0.010 \\
      \midrule

        \multicolumn{9}{c}{Running time (s)}\\
        \midrule
        GEE & 0.01$\pm$0.00 & 0.00$\pm$0.00 & 0.00$\pm$0.00 & 0.00$\pm$0.00 & 0.01$\pm$0.00 & 0.01$\pm$0.00 & 0.00$\pm$0.00 & 0.00$\pm$0.00 \\
        GNN & \textbf{0.37$\pm$0.01} & \textbf{0.14$\pm$0.00} & 0.30$\pm$0.01 & \textbf{0.21$\pm$0.01} & 0.32$\pm$0.01 & \textbf{0.74$\pm$0.02} & 0.16$\pm$0.00 & \textbf{0.13$\pm$0.00} \\
        GG/GG-C  & 0.37$\pm$0.01 & 0.14$\pm$0.00 & \textbf{0.28$\pm$0.01} & 0.22$\pm$0.01 & \textbf{0.30$\pm$0.01} & 0.77$\pm$0.02 & \textbf{0.14$\pm$0.00} & 0.13$\pm$0.00 \\
        \bottomrule
    \end{tabular}
  \end{subtable}

  \vspace{0.6em}

  \begin{subtable}[t]{\textwidth}
    \centering
    \label{tab:part2}
    
    \begin{tabular}{lcccccccc}
      \toprule
      5\% & LastFM & PolBlogs & TerroristRel & KarateClub & Chameleon & Cora & Citeseer \\
      \midrule
      \multicolumn{8}{c}{Accuracy}\\
      \midrule
      GEE & 0.333$\pm$0.004 & 0.723$\pm$0.005 & \textit{0.737$\pm$0.010} & 0.648$\pm$0.016 & 0.267$\pm$0.004 & 0.349$\pm$0.006 & 0.244$\pm$0.002 \\
      GNN & 0.431$\pm$0.003 & 0.706$\pm$0.009 & 0.683$\pm$0.006 & 0.257$\pm$0.012 & \textit{0.288$\pm$0.003} & 0.318$\pm$0.006 & 0.238$\pm$0.003 \\
      GG  & \textbf{0.536$\pm$0.003} & \textit{0.729$\pm$0.014} & 0.718$\pm$0.011 & \textit{0.663$\pm$0.019} & 0.270$\pm$0.004 & \textit{0.393$\pm$0.005} & \textit{0.280$\pm$0.003} \\
      GG-C & \textit{0.509$\pm$0.003} & \textbf{0.815$\pm$0.009} & \textbf{0.765$\pm$0.006} & \textbf{0.729$\pm$0.014} & \textbf{0.333$\pm$0.003} & \textbf{0.426$\pm$0.004} & \textbf{0.303$\pm$0.003} \\
      \midrule
        \multicolumn{8}{c}{Running time (s)}\\
        \midrule
        GEE & 0.01$\pm$0.00 & 0.01$\pm$0.00 & 0.00$\pm$0.00 & 0.00$\pm$0.00 & 0.01$\pm$0.00 & 0.00$\pm$0.00 & 0.00$\pm$0.00 \\
        GNN & \textbf{1.58$\pm$0.04} & 0.34$\pm$0.01 & 0.26$\pm$0.01 & 0.13$\pm$0.00 & 0.57$\pm$0.01 & \textbf{0.38$\pm$0.01} & \textbf{0.36$\pm$0.01} \\
        GG/GG-C  & 2.18$\pm$0.05 & \textbf{0.30$\pm$0.01} & \textbf{0.25$\pm$0.01} & \textbf{0.13$\pm$0.00} & \textbf{0.56$\pm$0.02} & 0.39$\pm$0.02 & 0.38$\pm$0.01 \\
        \bottomrule
    \end{tabular}
  \end{subtable}
  
\vspace{2.5em}
  \begin{subtable}[t]{\textwidth}
    \centering
    \begin{tabular}{lcccccccc}
      \toprule
      50\% & ACM & BAT & DBLP & EAT & UAT & Wiki & Gene & Industry \\
      \midrule
      \multicolumn{9}{c}{Accuracy}\\
      \midrule
      GEE & 0.749$\pm$0.002 & 0.350$\pm$0.008 & \textbf{0.538$\pm$0.001} & 0.314$\pm$0.008 & \textbf{0.497$\pm$0.003} & 0.592$\pm$0.001 & \textbf{0.762$\pm$0.002} & 0.601$\pm$0.005 \\
      GNN & 0.544$\pm$0.008 & \textit{0.370$\pm$0.007} & 0.379$\pm$0.005 & \textit{0.357$\pm$0.005} & 0.385$\pm$0.006 & 0.406$\pm$0.006 & 0.606$\pm$0.009 & \textit{0.511$\pm$0.006} \\
      GG  & \textit{0.751$\pm$0.004} & 0.319$\pm$0.008 & \textit{0.532$\pm$0.002} & 0.248$\pm$0.005 & 0.439$\pm$0.005 & \textbf{0.617$\pm$0.001} & 0.704$\pm$0.009 & \textit{0.614$\pm$0.005} \\
      GG-C & \textbf{0.759$\pm$0.002} & \textbf{0.394$\pm$0.007} & 0.536$\pm$0.002 & \textbf{0.385$\pm$0.007} & \textit{0.472$\pm$0.004} & \textit{0.612$\pm$0.001} & \textit{0.744$\pm$0.003} & \textbf{0.624$\pm$0.005} \\
      \midrule
      \multicolumn{9}{c}{Running time (s)}\\
      \midrule
      GEE      & 0.01$\pm$0.00 & 0.00$\pm$0.00 & 0.00$\pm$0.00 & 0.00$\pm$0.00 & 0.01$\pm$0.00 & 0.01$\pm$0.00 & 0.00$\pm$0.00 & 0.00$\pm$0.00 \\
      GNN      & \textbf{0.40$\pm$0.01} & 0.14$\pm$0.00 & \textbf{0.32$\pm$0.01} & 0.24$\pm$0.01 & \textbf{0.37$\pm$0.01} & \textbf{0.85$\pm$0.04} & 0.19$\pm$0.01 & 0.14$\pm$0.00 \\
      GG/GG-C   & 0.70$\pm$0.03 & \textbf{0.13$\pm$0.00} & 0.54$\pm$0.02 & \textbf{0.21$\pm$0.01} & 0.47$\pm$0.02 & 0.93$\pm$0.02 & \textbf{0.18$\pm$0.01} & \textbf{0.14$\pm$0.00} \\
      \bottomrule
    \end{tabular}
  \end{subtable}

  \vspace{0.6em}

  \begin{subtable}[t]{\textwidth}
    \centering
    \begin{tabular}{lccccccc}
      \toprule
      50\% & LastFM & PolBlogs & TerroristRel & KarateClub & Chameleon & Cora & Citeseer \\
      \midrule
      \multicolumn{8}{c}{Accuracy}\\
      \midrule
      GEE & 0.717$\pm$0.001 & \textit{0.907$\pm$0.002} & \textit{0.899$\pm$0.001} & \textit{0.826$\pm$0.011} & 0.263$\pm$0.004 & 0.724$\pm$0.001 & 0.545$\pm$0.001 \\
      GNN & 0.621$\pm$0.003 & 0.737$\pm$0.012 & 0.712$\pm$0.012 & 0.390$\pm$0.015 & \textit{0.281$\pm$0.002} & 0.477$\pm$0.006 & 0.373$\pm$0.005 \\
      GG  & \textbf{0.773$\pm$0.001} & 0.869$\pm$0.011 & 0.865$\pm$0.010 & 0.803$\pm$0.014 & 0.320$\pm$0.004 & \textit{0.739$\pm$0.002} & \textit{0.577$\pm$0.002} \\
      GG-C & \textit{0.766$\pm$0.001} & \textbf{0.915$\pm$0.002} & \textbf{0.908$\pm$0.002} & \textbf{0.838$\pm$0.009} & \textbf{0.379$\pm$0.004} & \textbf{0.749$\pm$0.002} & \textbf{0.582$\pm$0.001} \\
      \midrule
      \multicolumn{8}{c}{Running time (s)}\\
      \midrule
      GEE      & 0.02$\pm$0.00 & 0.01$\pm$0.00 & 0.01$\pm$0.00 & 0.00$\pm$0.00 & 0.02$\pm$0.00 & 0.00$\pm$0.00 & 0.00$\pm$0.00 \\
      GNN      & \textbf{1.58$\pm$0.04} & \textbf{0.43$\pm$0.01} & \textbf{0.29$\pm$0.01} & \textbf{0.12$\pm$0.00} & \textbf{0.62$\pm$0.03} & \textbf{0.46$\pm$0.02} & \textbf{0.44$\pm$0.01} \\
      GG/GG-C   & 2.80$\pm$0.08 & 0.48$\pm$0.03 & 0.36$\pm$0.02 & 0.12$\pm$0.00 & 1.05$\pm$0.05 & 0.78$\pm$0.03 & 0.74$\pm$0.02 \\
      \bottomrule
    \end{tabular}
  \end{subtable}
\vspace{-2em}
\end{table*}

\section{Classification Evaluation} \label{sec: Classification Evaluation}

In this section, we evaluate the node classification performance of the proposed methods. The evaluation is conducted on both synthetic graphs generated by the SBM and DC-SBM, as well as on the same set of real-world datasets used in the clustering analysis. A key part of this evaluation is to assess model performance under varying levels of data scarcity, which is achieved through different training sizes.

\subsection{Experimental Setup} \label{sec: Classification Experimental Setup}
We systematically vary the proportion of labeled nodes available for training by four different training sizes. Specifically, we set the ratio of the test set ($\mathcal{V}_{\text{test}}$) to the combined training and validation sets (train/val set) to be 19:1, 9:1, 4:1, and 1:1. This corresponds to the use of 5\%, 10\%, 20\%, and 50\% of the nodes for training and validation, respectively. Furthermore, to ensure that every class is represented during training and validation, we enforce a minimum of two training nodes and one validation node per class. 

From this combined training and validation pool, we further allocate 10\% of the nodes for the validation set ($\mathcal{V}_{\text{val}}$) and the remaining 90\% for the training set ($\mathcal{V}_{\text{train}}$). 

Each method is evaluated by the classification accuracy. For a given set of nodes $\mathcal{V}$, accuracy is defined as the fraction of correctly classified nodes. We denote this metric as $\text{Acc}(\mathcal{V})$ and formalize it as:
\begin{align*}
    \text{Acc}(\mathcal{V}) = \frac{1}{|\mathcal{V}|} \sum_{i \in \mathcal{V}} \mathbb{I}(\hat{Y}_i = Y_i),
    \label{eq:accuracy}
\end{align*}
where $\hat{Y}_i$ is the predicted label for node $i$ and $Y_i$ is its ground-truth label. 

During training, we employ an early stopping strategy for the GNN-based methods by monitoring accuracy on a separate validation set, $\text{Acc}(\mathcal{V}_{\text{val}})$. The model parameters from the epoch that yields the highest validation accuracy are restored for the final evaluation on the test set, $\text{Acc}(\mathcal{V}_{\text{test}})$. To ensure a fair comparison, GEE is also evaluated under the same conditions, where the labels of nodes in the validation set are masked alongside the test set labels and are not used during its embedding generation process.

Regarding the running time of GG-C, it is reported jointly with GG in our figures and tables. This is because the additional computation in GG-C, consisting only of a simple feature concatenation and an LDA step, is negligible compared to the shared GNN training process. As a result, the running times of GG and GG-C are almost always identical.



\subsection{Simulation} \label{sec: Classification Simulation}

Here we present two most representative train/val set proportions on DC-SBM: 50\% (data-rich) and 5\% (data-scarce), in Figure~\ref{fig:classification DC_SBM 5 and 50}. The complete results, including those for the 10\% and 20\% train/val set proportions on DC-SBM and all results for SBM, are provided in Appendix. The graph generation parameters for this classification task are identical to those used in the clustering evaluation, except that we perform 200 independent runs.

From the left two panels, it is evident that both GG and GG-C significantly outperform the vanilla GNN. When provided with sufficient training data (e.g., $50\%$ of nodes labeled), GEE performs remarkably well, aligning with its theoretical guarantees. In such cases, GG and GG-C achieve similar high accuracy, indicating that GEE alone is already near-optimal. As the proportion of training/validation data decreases (see Appendix for results under $10\%$ and $20\%$ training), GEE's advantage decreases, though it still remains superior to the vanilla GNN.

In the semi-supervised regime, GG provides meaningful refinement over GEE, but its standalone accuracy does not substantially exceed GEE’s until GG-C is introduced. This reveals an important insight: GEE supplies a strong global initialization, GG contributes nontrivial local refinement, and their concatenation yields a fused representation that effectively synergizes GEE’s global structural information with the GNN’s localized, task-specific adjustments.

From the right two panels, an interesting observation emerges: the vanilla GNN appears to converge slightly faster than the GG-based models. However, this should not be interpreted as better efficiency. Rather, it reflects early termination to a suboptimal local minimum. Note that the time difference between GG and vanilla GNN is marginal and negligible across all classification tasks.

\subsection{Real data} \label{sec: Classification Real Data}

For experiments on real-world datasets, we performed 100 independent runs for each data split to ensure statistical robustness. All publicly available datasets were included in our evaluation, with the exception of the Email dataset, which was excluded due to several classes containing fewer than three samples, violating our minimum class size requirement. Results for the 5\% and 50\% training/validation set proportions are presented in Table~\ref{tab:classification realdata 5 and 50}. Additional results for the 10\% and 20\% settings are provided in Appendix.

Both GG and GEE serve as strong baselines, and in most cases outperform the vanilla GNN. Neither consistently dominates the other across all datasets. By contrast, the GG-C version, which fuses the embeddings from both GG and GEE, successfully captures complementary information from both global and local representations. As a result, it achieves superior performance across a diverse set of graph data. In nearly every case, GG-C either ranks as the top performer or closely follows the best-performing method in terms of classification accuracy. Finally, an expanded set of evaluations is provided in appendix, including a comparison against the recent CutSSL~\cite{holtz2024continuous} baseline and an analysis of our models' performance with increased layer depth.

\section{Conclusion} 

In this work, we introduced the GEE-powered GNN framework, motivated by the hypothesis that informative node features can substantially improve upon random initializations in graph neural networks, alleviating issues such as slow convergence and suboptimal solutions. Through extensive experiments on both synthetic and real-world datasets, we demonstrated that the proposed method consistently enhances performance across node-level tasks. For node clustering, the GEE-based warm-start leads to faster convergence and superior clustering quality. For node classification, our concatenated variant achieves excellent performance by retaining the global information captured by GEE while benefiting from the local refinements introduced by the GNN.

These findings point to a promising direction in GNN research: rather than focusing solely on deeper or more sophisticated neural architectures, substantial improvements can be achieved by adopting high-quality feature initialization. Such strategies act as a form of implicit regularization, improving generalization and reducing the risk of overfitting, particularly in low-label or semi-supervised settings. Among the many available graph embedding algorithms, we chose GEE for its statistical consistency and computational scalability, but other initialization strategies could certainly be explored in future work.

The broader implications of this work extend beyond graph learning. The idea of leveraging statistically informed, structure-aware initializations is applicable to a wide range of neural network architectures, both for graph-structured and conventional data. Several promising directions remain, including exploring alternative warm-start strategies and deeper architectures; extending the approach to dynamic or temporal graphs, where initializations can adapt to evolving structure; and investigating applications in domains such as recommendation systems or computer vision, where combining pretrained global features with local refinement may offer similar advantages.

\bibliographystyle{IEEEtran} 
\bibliography{referencesMain} 

\begin{IEEEbiography}
[{\includegraphics[width=1in,height=1.25in,clip,keepaspectratio]{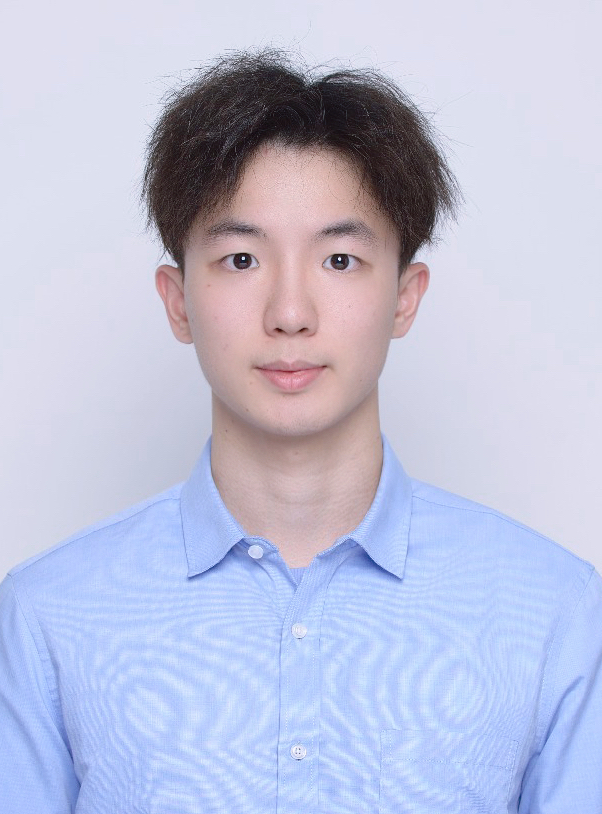}}]{Shiyu Chen received the B.E. degree in computer science from Sun Yat-sen University, Guangzhou, China, in 2024, and the M.S. degree in data science from Johns Hopkins University in 2025. His research interests include graph neural networks, random graph models, and statistical learning on graphs.}
\end{IEEEbiography}
\begin{IEEEbiography}
[{\includegraphics[width=1in,height=1.25in,clip,keepaspectratio]{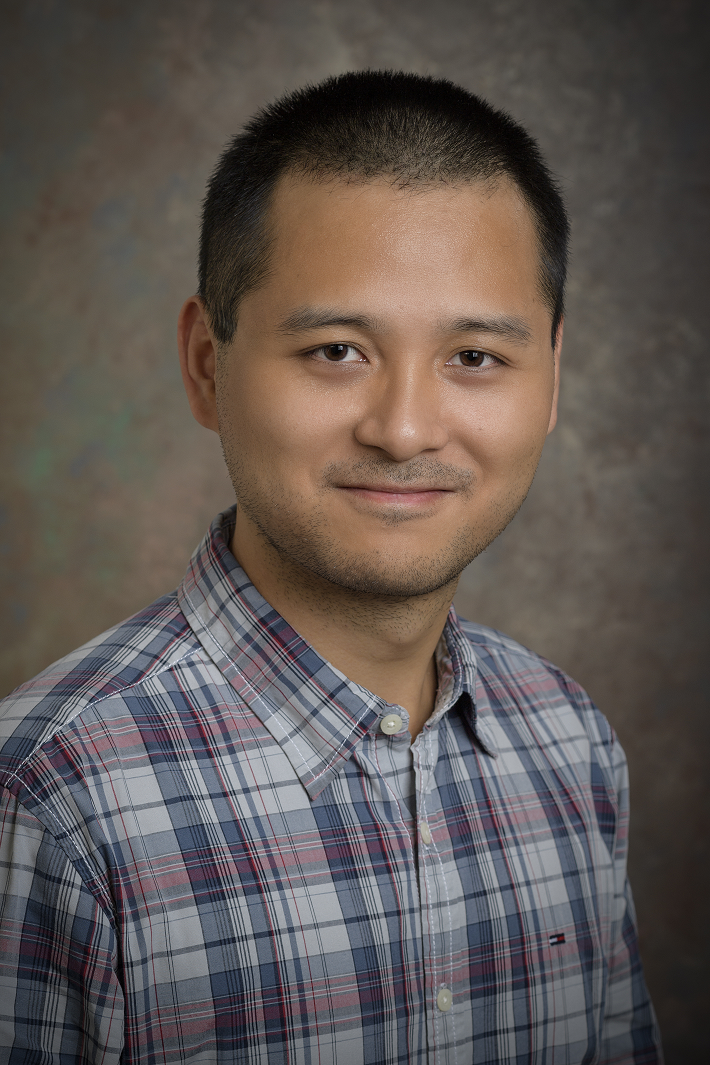}}]{Cencheng Shen is a Principal Research Data Scientist at Microsoft Research. He received the BS degree in Quantitative Finance from the National University of Singapore, and the PhD degree in Applied Mathematics and Statistics from Johns Hopkins University. Prior to joining Microsoft Research, he was an Associate Professor in the Department of Applied Economics and Statistics at the University of Delaware.}
\end{IEEEbiography}
\begin{IEEEbiography}
[{\includegraphics[width=1in,height=1.25in,clip,keepaspectratio]{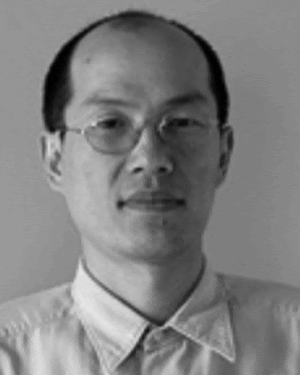}}]{Youngser Park received the B.E. degree in electrical engineering from Inha University in Seoul, Korea in 1985, the M.S. and Ph.D. degrees in computer science from The George Washington University in 1991 and 2011 respectively. From 1998 to 2000 he worked at the Johns Hopkins Medical Institutes as a senior research engineer. From 2003 until 2011 he worked as a senior research analyst, and has been an associate research scientist since 2011 then research scientist since 2019 in the Center for Imaging Science at the Johns Hopkins University. At Johns Hopkins, he holds joint appointments in the Mathematical Institute for Data Science and the Human Language Technology Center of Excellence. His current research interests are clustering algorithms, pattern classification, and data mining for high-dimensional and graph data.}
\end{IEEEbiography}
\begin{IEEEbiography}
[{\includegraphics[width=1in,height=1.25in,clip,keepaspectratio]{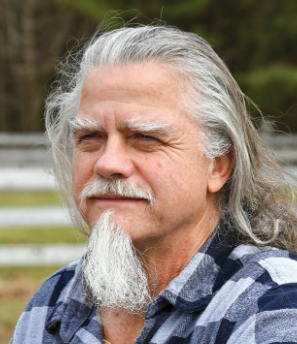}}]{Carey E. Priebe received the BS degree in mathematics from Purdue University in 1984, the MS degree in computer science from San Diego State University in 1988, and the PhD degree in information technology (computational statistics) from George Mason University in 1993. From 1985 to 1994 he worked as a mathematician and scientist in the US Navy research and development laboratory system. Since 1994 he has been a professor in the Department of Applied Mathematics and Statistics at Johns Hopkins University. His research interests include computational statistics, kernel and mixture estimates, statistical pattern recognition, model selection, and statistical inference for high-dimensional and graph data. He is a Senior Member of the IEEE, an Elected Member of the International Statistical Institute, a Fellow of the Institute of Mathematical Statistics, and a Fellow of the American Statistical Association.}
\end{IEEEbiography}

\if0\blind
{
 \clearpage
  \onecolumn

\setcounter{figure}{0}
\renewcommand{\thealgorithm}{C\arabic{algorithm}}
\renewcommand{\thefigure}{E\arabic{figure}}
\renewcommand{\thesubsection}{\thesection.\arabic{subsection}}
\renewcommand{\thesubsubsection}{\thesubsection.\arabic{subsubsection}}

\bigskip
\begin{center}
{\large\bf APPENDIX}
\end{center}

\section{Implementation Details}
For the clustering task, we used the Adam optimizer with a learning rate of 0.01. Models were trained for a maximum of 10,000 epochs, with early stopping based on the Adjusted Rand Index (ARI) score, using a patience of 800 epochs. For the node classification task, the model was trained with a Cross-Entropy loss. We used the Adam optimizer with a learning rate of 0.001 and a weight decay of 5e-4. Training ran for up to 10,000 epochs, with early stopping determined by validation accuracy, using a patience of 100 epochs. All experiments were conducted on a MacBook equipped with an Apple M2 chip and 16 GB of RAM. 

 \section{Clustering Evaluation} \label{Appendix: Clustering Evaluation}

 This section provides supplementary results for our methods on graphs generated by Stochastic Block Model (SBM)~\cite{holland1983stochastic}. The SBM can be understood as a simplified case of the DC-SBM (described in the main text) that does not account for degree heterogeneity. The probability of an edge between nodes $i$ and $j$ depends solely on their community assignments, with the adjacency matrix $A$ being generated as: $A(i,j) \sim \text{Bernoulli}(B(Y_i, Y_j))$.

 For this experiment, we generate graphs with $n=800$ nodes and $K=4$ imbalanced communities (1:2:3:4 ratio). We fix the intra-community link probability to $0.15$ and vary the inter-community link probability, $r$, from $0$ to $0.15$.

 The results for clustering accuracy (ARI) and running time are presented in Figure~\ref{fig:clustering-SBM}. The trends are highly consistent with the DC-SBM experiments, confirming that GG is the best overall method. While GEE excels in low-noise settings ($r<0.06$), its performance sharply declines in more challenging scenarios. This dropoff is so significant that it even hinders GG's performance, which underscores the need for a high-quality and stable initialization to achieve optimal results efficiently.

 \section{Classification Evaluation} \label{Appendix: Classification Evaluation}

 \subsection{Simulation} \label{Appendix: Classification Simulation}
 This subsection provides supplementary results for our simulation studies on classification task. To offer a more complete picture of model performance on DC-SBM under varying data scarcity, we present the results for training proportions of 10\% and 20\% in Figure~\ref{fig:classification DC_SBM 10 and 20}. 

 Additionally, for the classification task on SBM graphs, we used the same graph generation parameters as in the clustering study but conducted 200 independent trials for each graph generated at every value of $r$ to ensure high statistical confidence. The results for SBM across all data-splitting ratios are presented in Figures~\ref{fig:classification SBM all}.

 \subsection{Real Data} \label{Appendix: Classification Real Data}
 Table~\ref{tab:classification realdata 10 and 20} summarizes the performance on real-world datasets with 10\% and 20\% of nodes used for training and validation. Across these settings, our GG-C consistently delivers the top performance, closely followed by our GG.

\subsection{Extended Analysis}
We conduct a more comprehensive set of experiments, with the results summarized in Table~\ref{tab:additional classification realdata}. We first compare our method against a recent baseline, CutSSL. Note that CutSSL is a transductive algorithm that performs global optimization using the entire graph, including test nodes, during training. 

The results show that CutSSL is a competitive method, performing better on some datasets and worse on others, an outcome that is reasonable given its fundamentally different algorithmic design. A key advantage of our approach, as well as of GEE and GNN, is that they are substantially faster than CutSSL. For example, on the LastFM dataset, CutSSL requires approximately 53 seconds, whereas our GG and GG-C models complete in just over 2 seconds. This highlights the strength of our framework in achieving an excellent balance between high accuracy and practical efficiency. Nevertheless, CutSSL’s accuracy advantage on certain datasets also highlights a complementary strength that we plan to further explore in future work.

We also examine the effect of increasing the number of GNN layers. Adding an additional layer (i.e., moving from the default 2-layer to a 3-layer architecture) generally yields a modest performance improvement. This supports our hypothesis that better initialization enhances deeper GNNs as well, suggesting that further gains may be possible by exploring more complex architectures combined with GEE initialization.

 \begin{figure*}[t] 
     \centering
     {
         \includegraphics[width=0.48\textwidth]{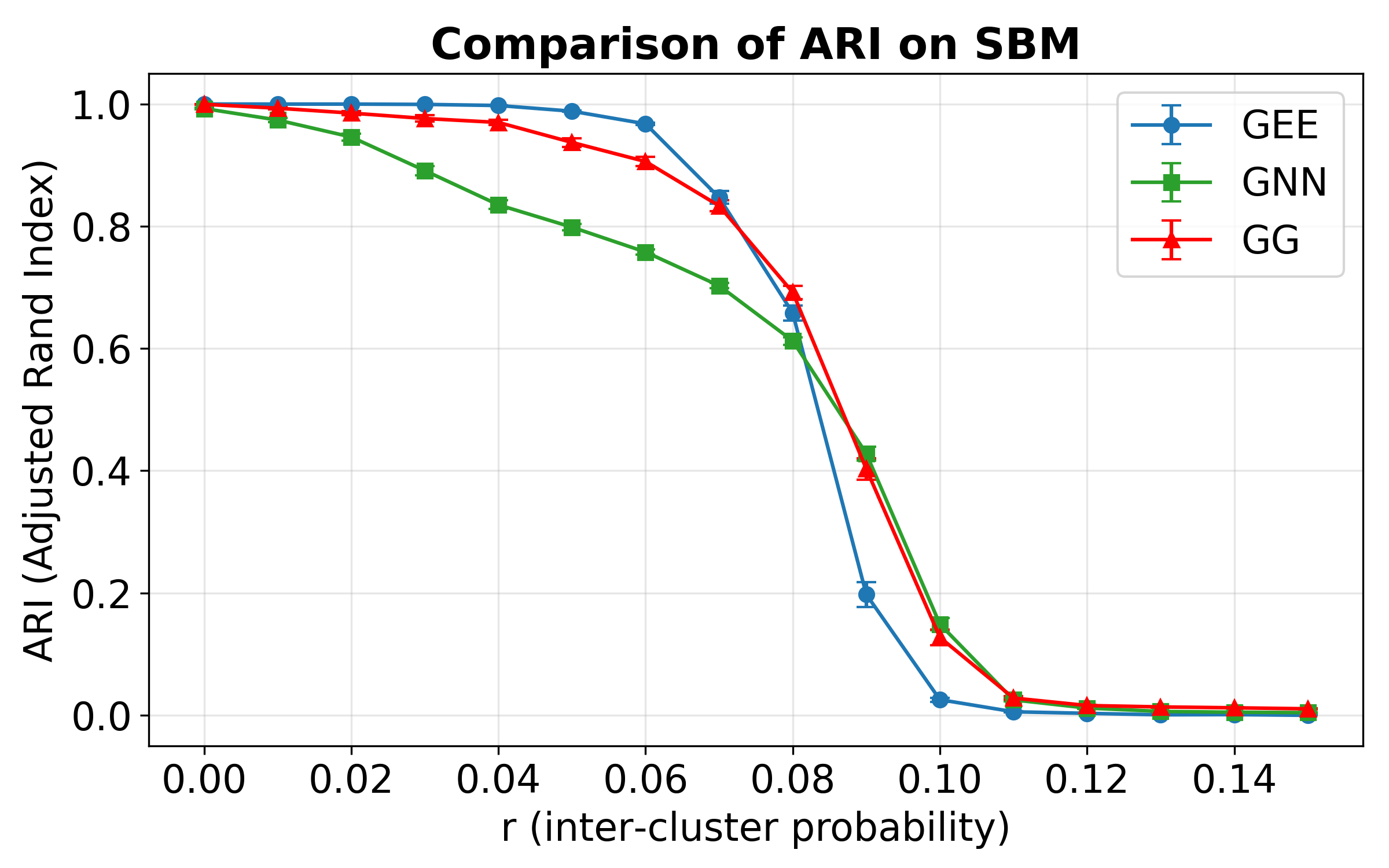}
     }
     \hfill
     {
         \includegraphics[width=0.48\textwidth]{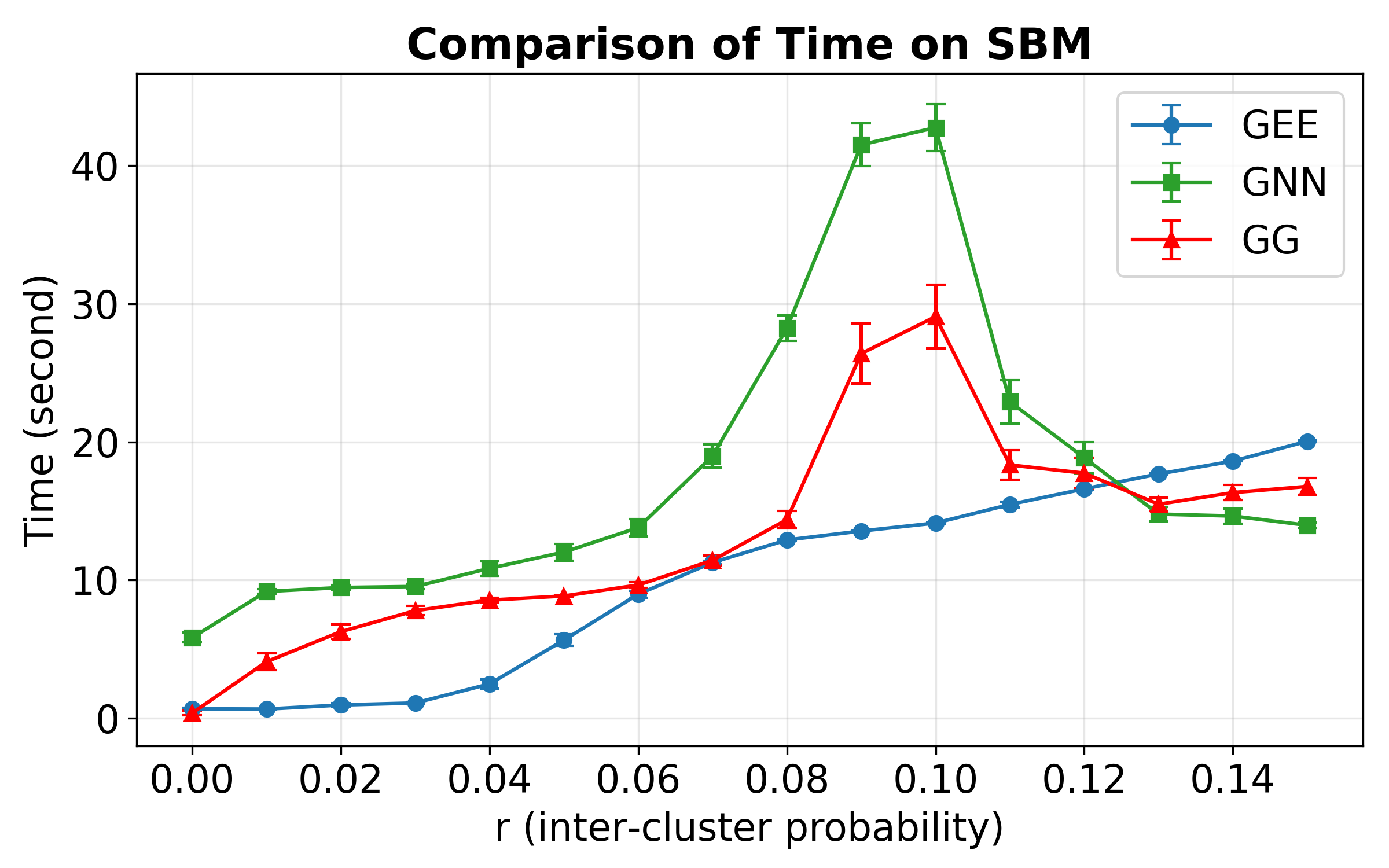}
     }
     \caption{Clustering performance comparison of GEE, GNN and GG on SBM graphs. This figure shows our GG method significantly outperforms the vanilla GNN in both accuracy and speed. Crucially, it also surpasses the classic GEE on more challenging graphs ($r > 0.08$), making it the most effective method overall.}
     \label{fig:clustering-SBM}
 \end{figure*}

 \begin{figure*}[t] 
     \centering
     {
         \includegraphics[width=0.48\textwidth]{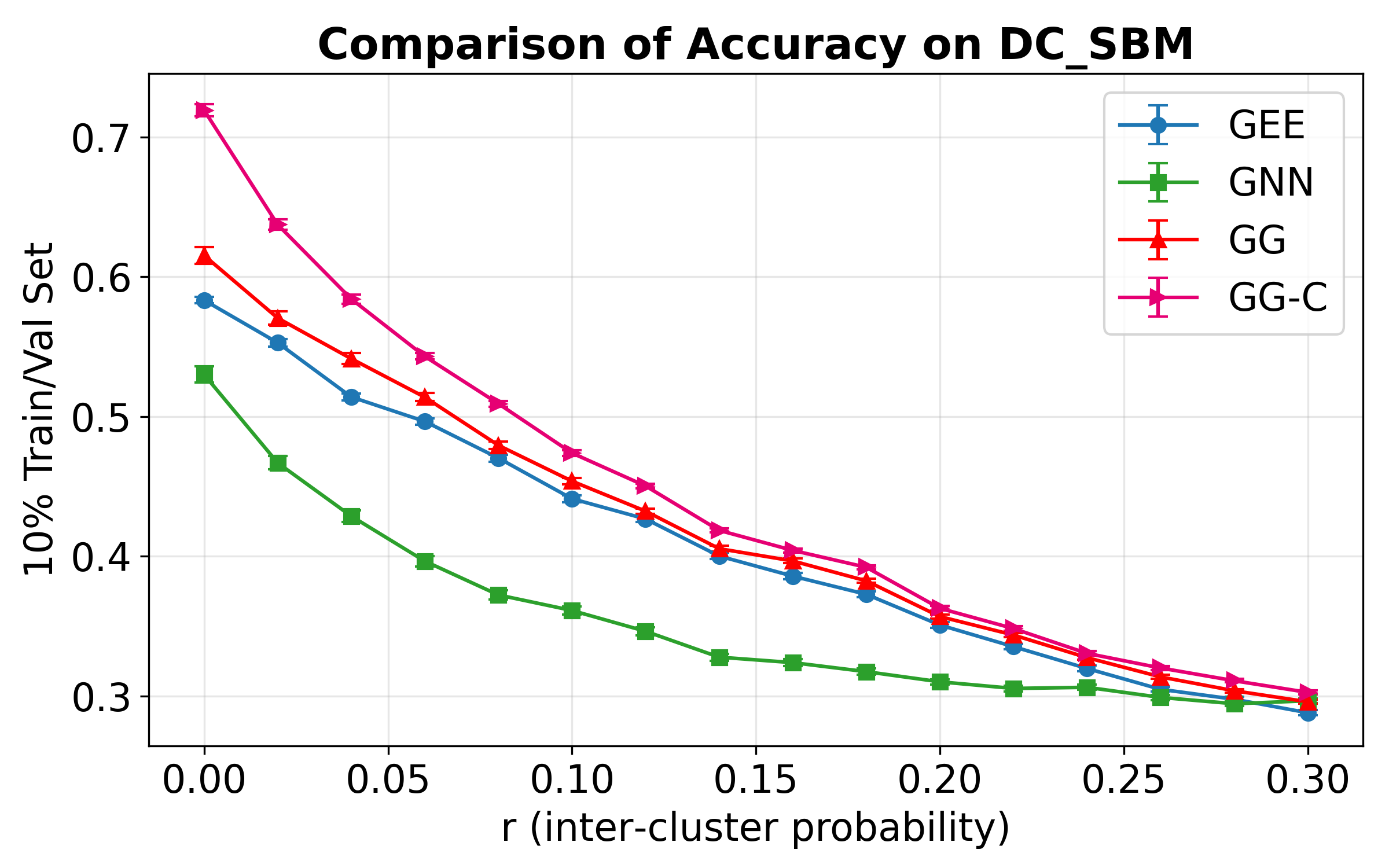}
         \label{fig:DC_SBM_acc_10}
     }
     \hfill
     {
         \includegraphics[width=0.48\textwidth]{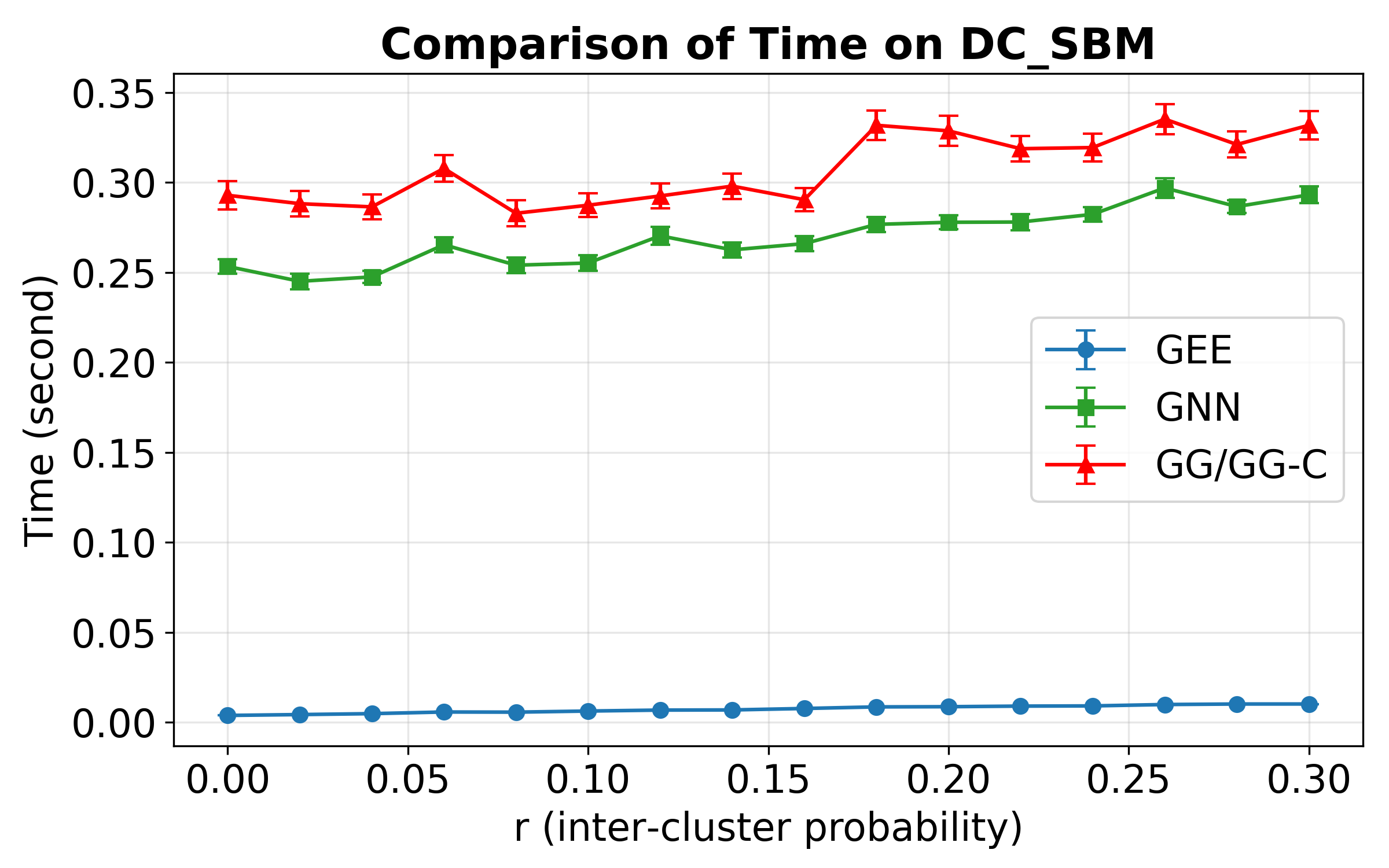}
         \label{fig:DC_SBM_time_10}
     }
     \centering
     {
         \includegraphics[width=0.48\textwidth]{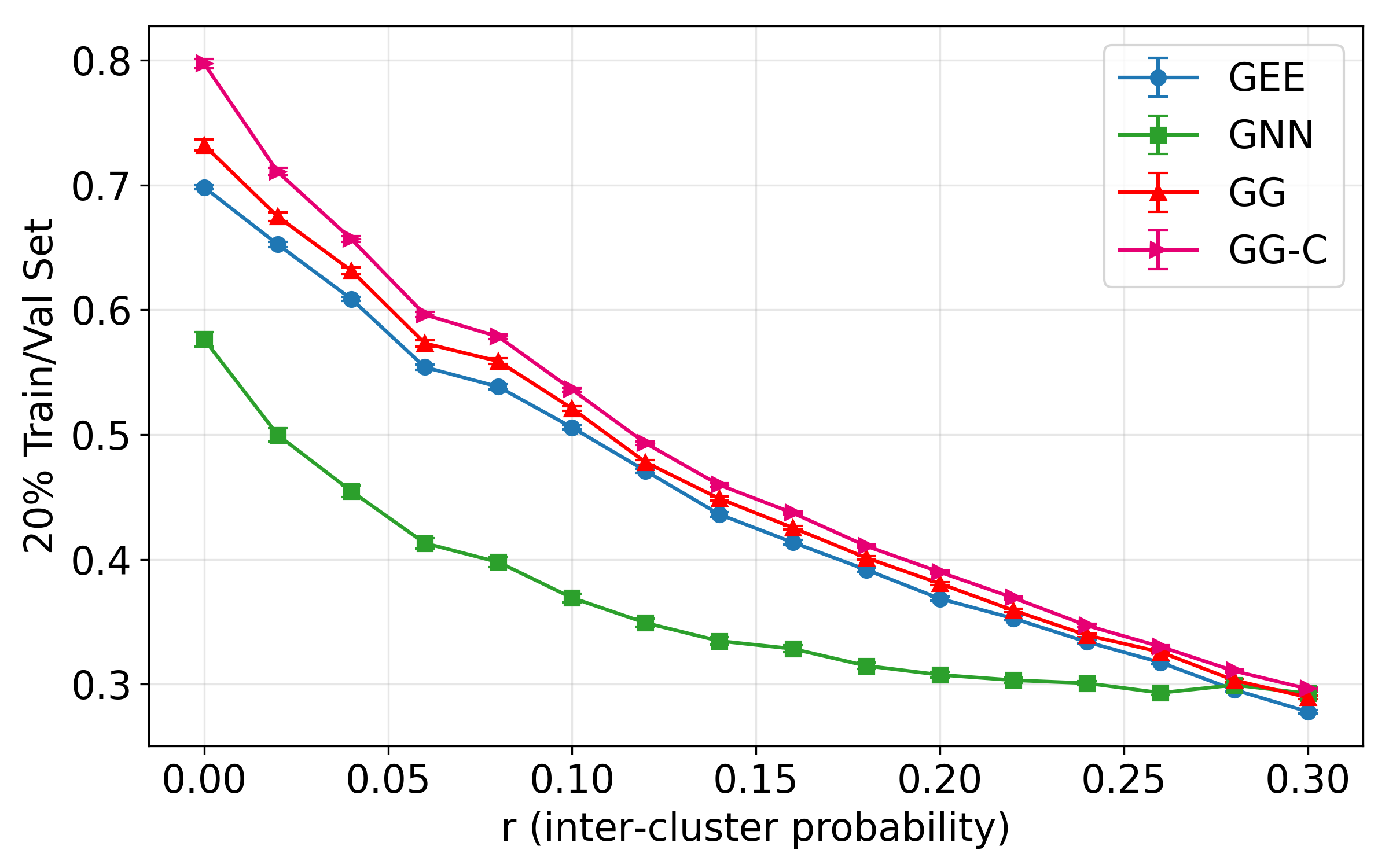}
         \label{fig:DC_SBM_acc_20}
     }
     \hfill
     {
         \includegraphics[width=0.48\textwidth]{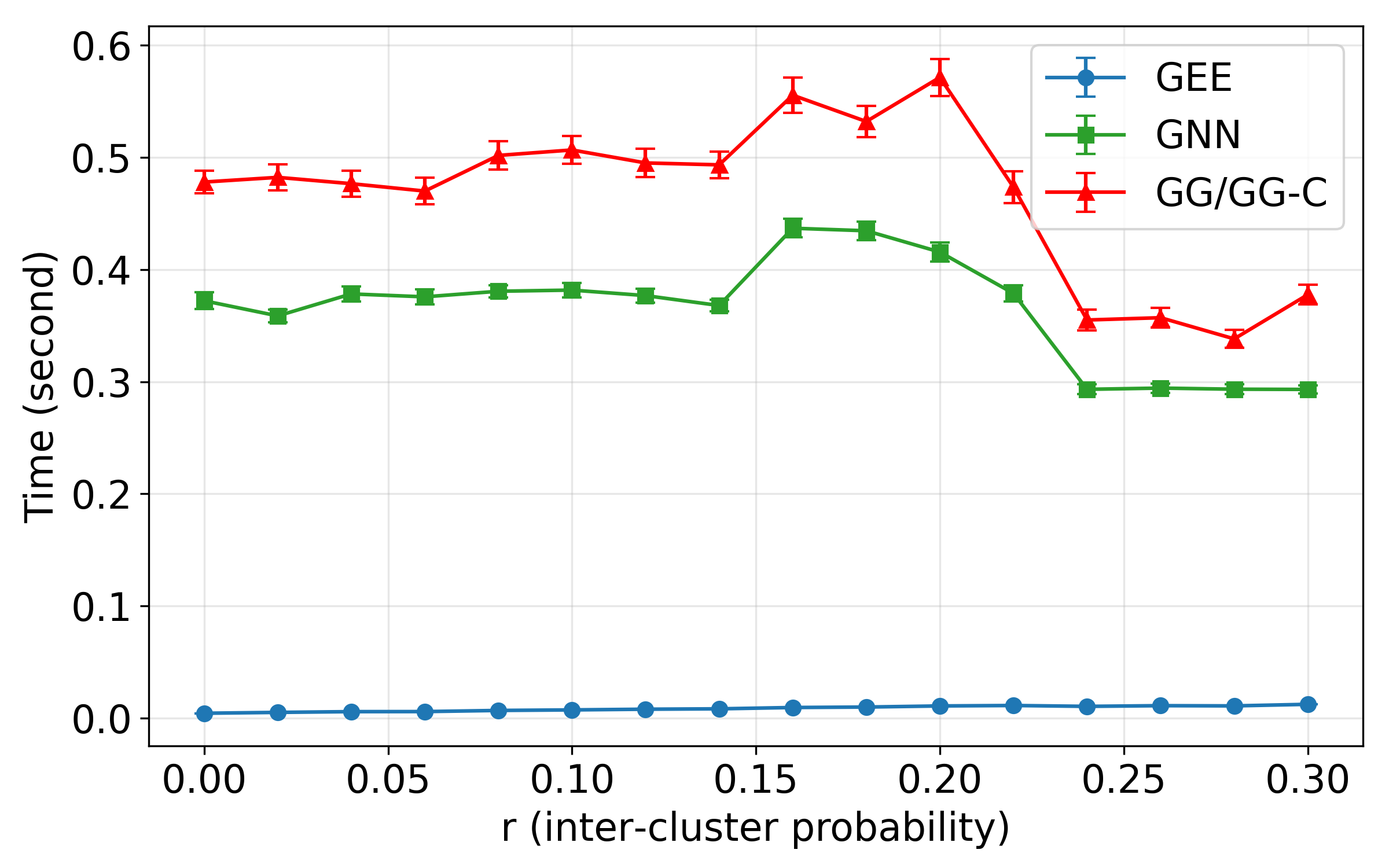}
         \label{fig:DC_SBM_time_20}
     }
     \caption{Classification performance comparison of GEE, GNN, GG and GG-C on DC-SBM graphs with 10\% (top row) and 20\% (bottom row) train/val set. The results show that GG and GG-C consistently deliver the best performance. They not only overcome the poor accuracy of the faster GNN, which is prone to local optima, but they also demonstrate a distinct advantage over GEE, especially as the amount of labeled data decreases.}
     \label{fig:classification DC_SBM 10 and 20}
      \vspace{-10pt}
 \end{figure*}

 \begin{figure*}[htbp] 
     \centering
     {
         \includegraphics[width=0.48\textwidth]{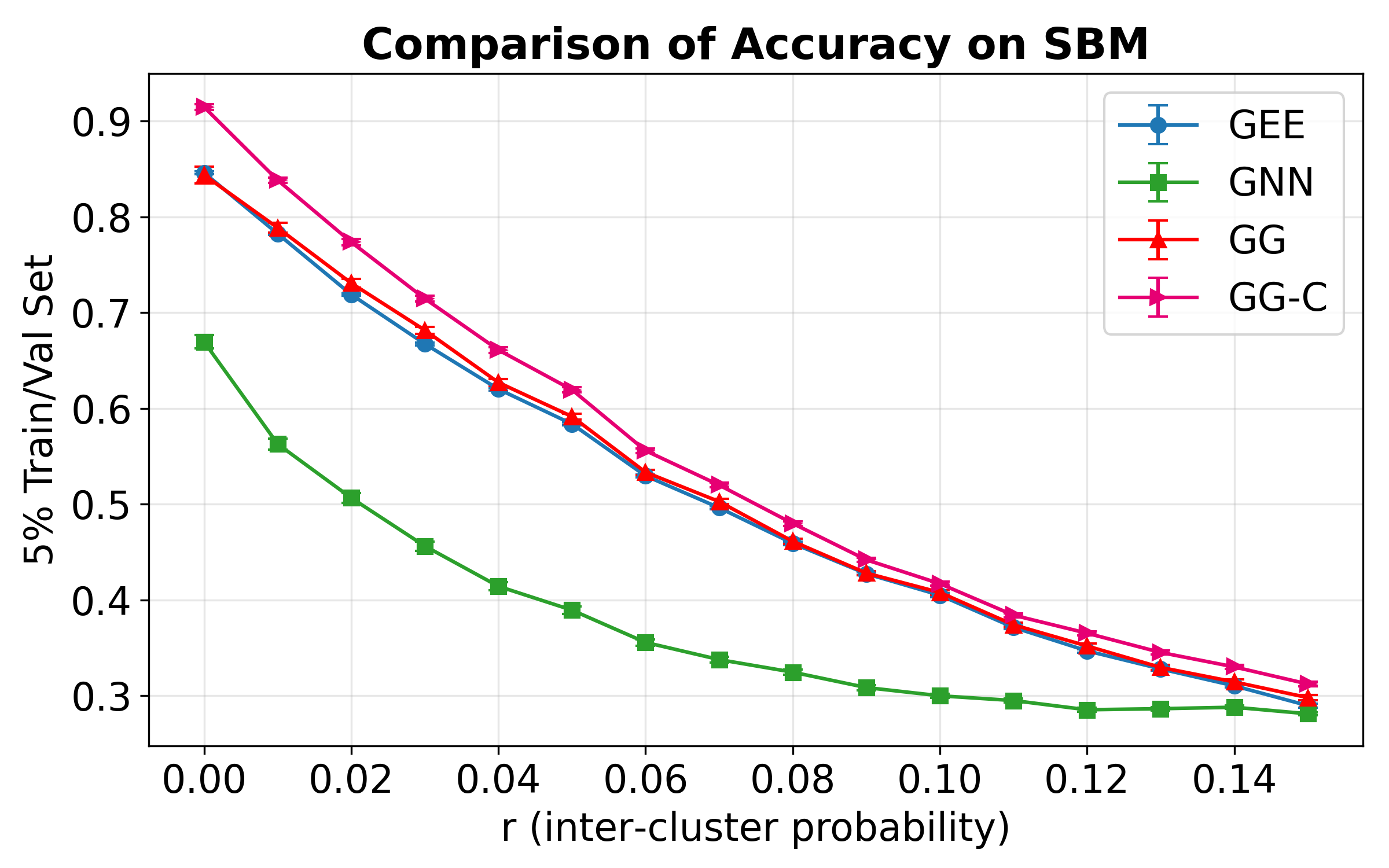}
         \label{fig:SBM_acc_5}
     }
     \hfill
     {
         \includegraphics[width=0.48\textwidth]{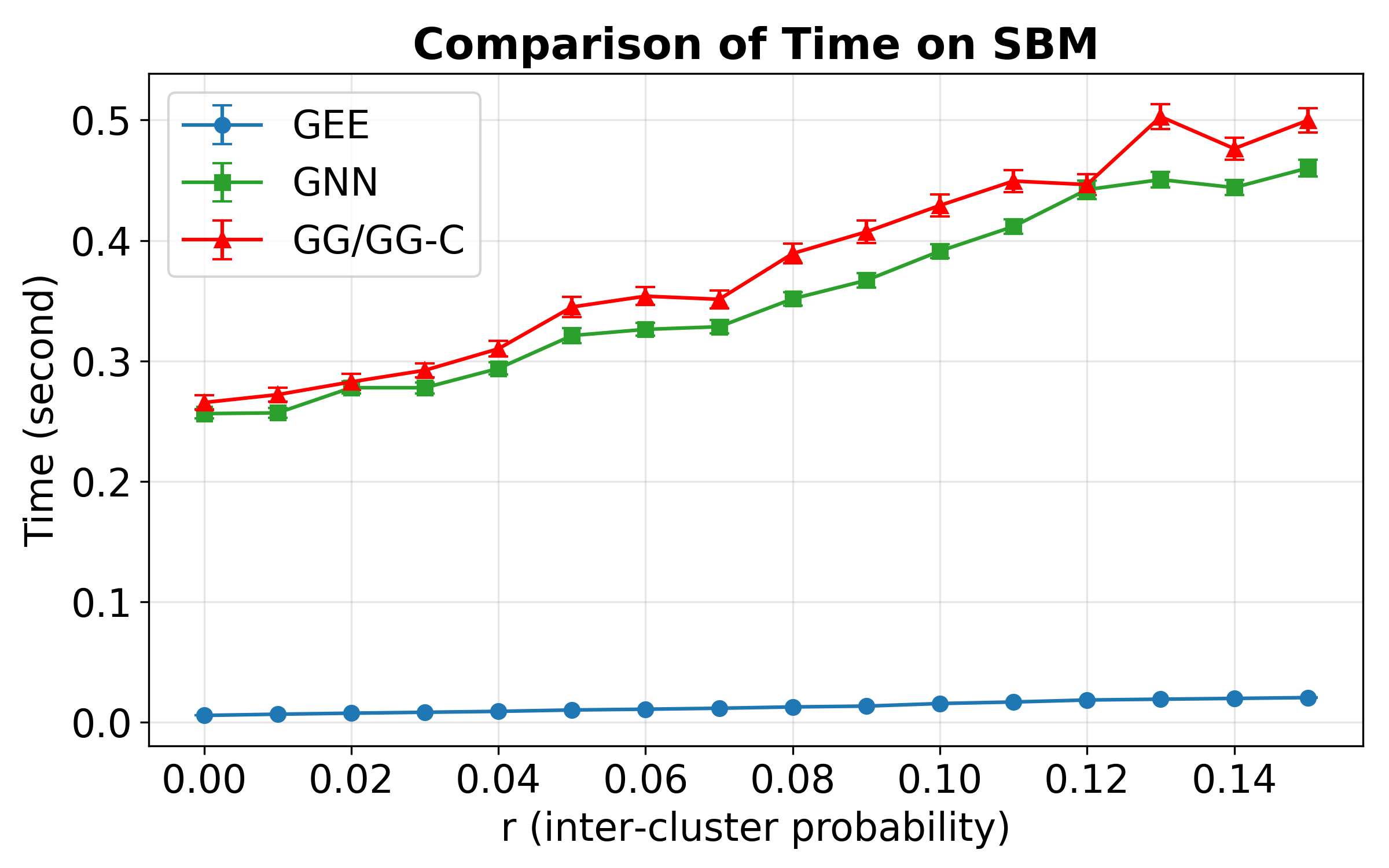}
         \label{fig:SBM_time_5}
     }
     \centering
     {
         \includegraphics[width=0.48\textwidth]{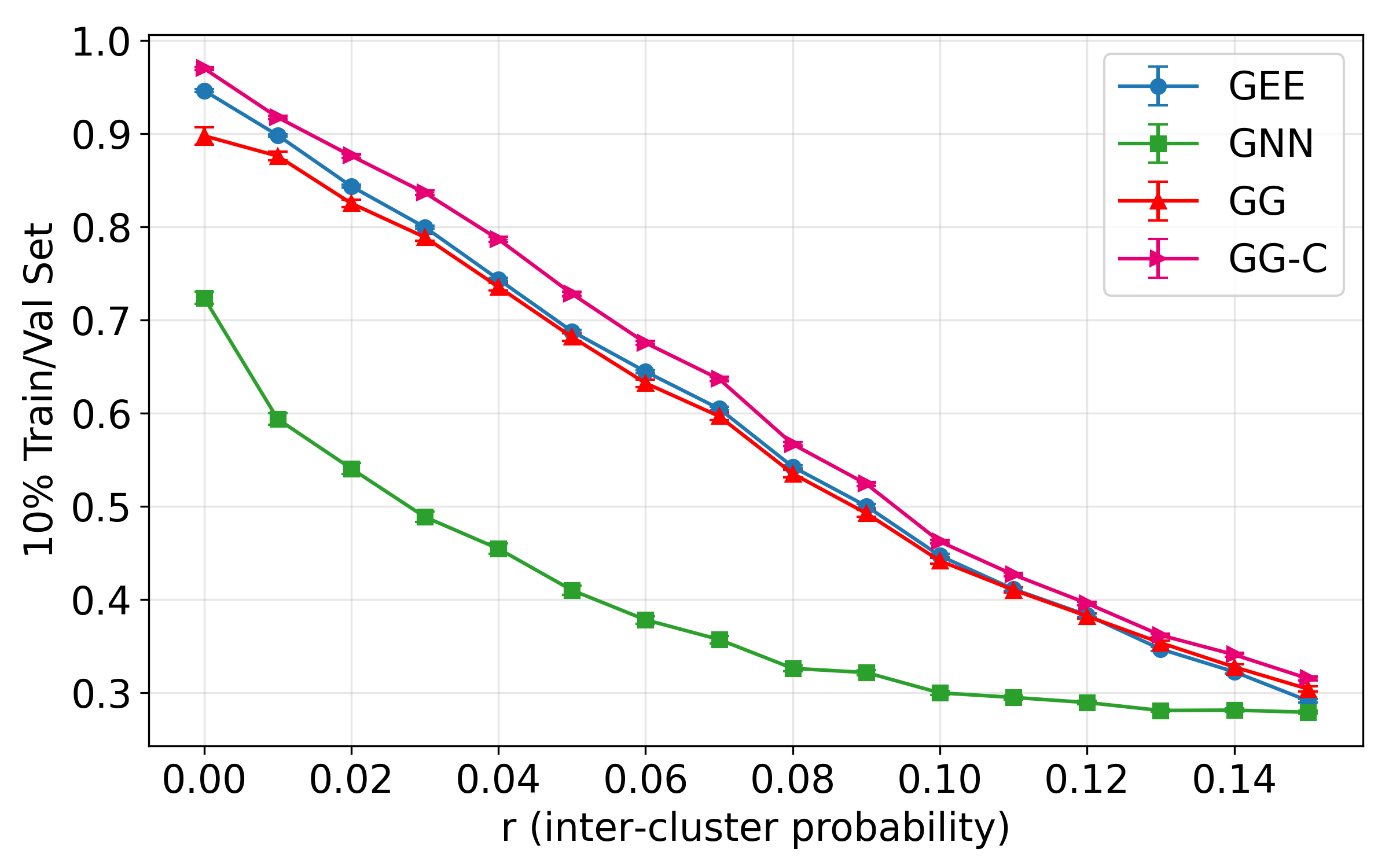}
         \label{fig:SBM_acc_10}
     }
     \hfill
     {
         \includegraphics[width=0.48\textwidth]{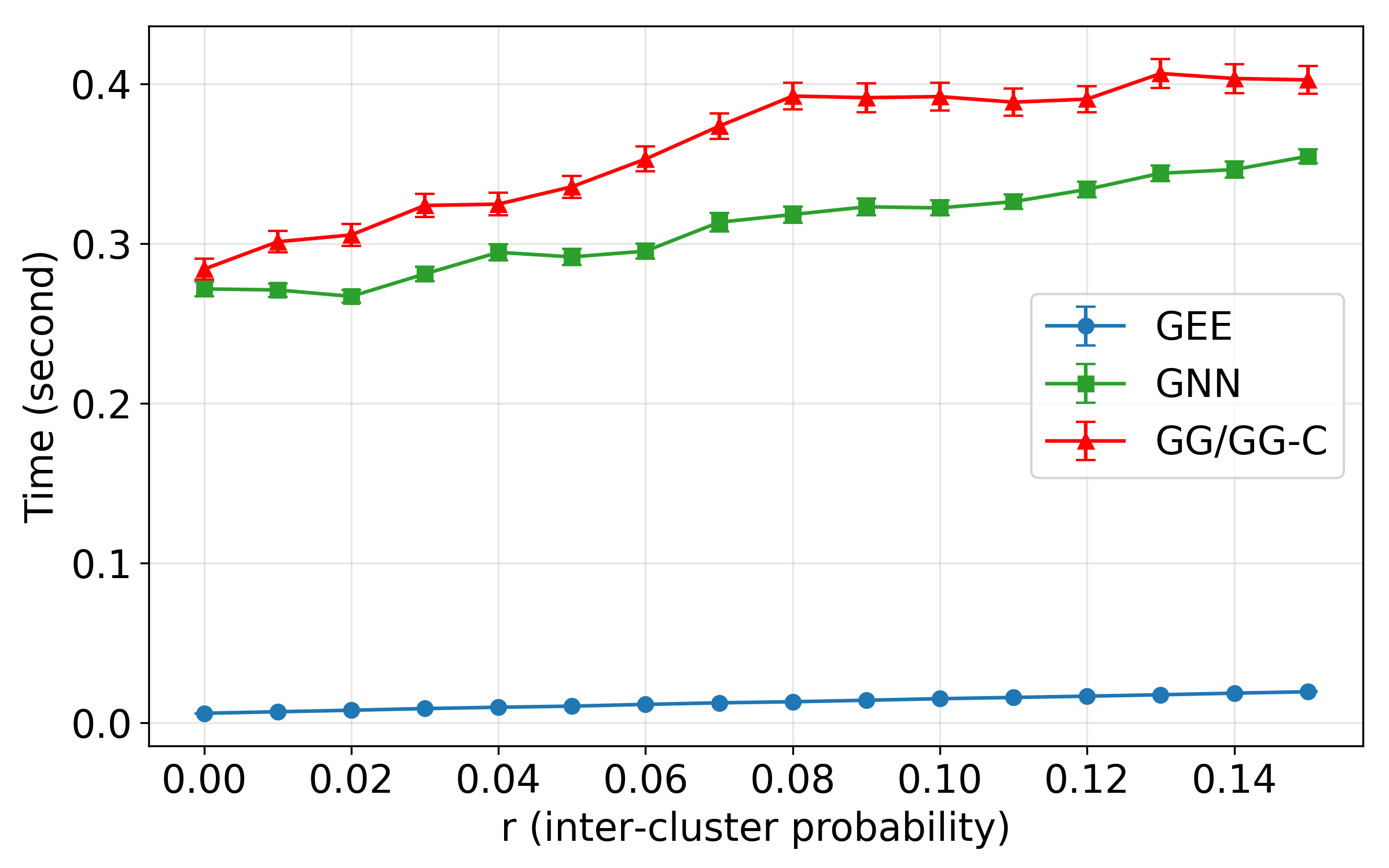}
         \label{fig:SBM_time_10}
     }
     \centering
     {
         \includegraphics[width=0.48\textwidth]{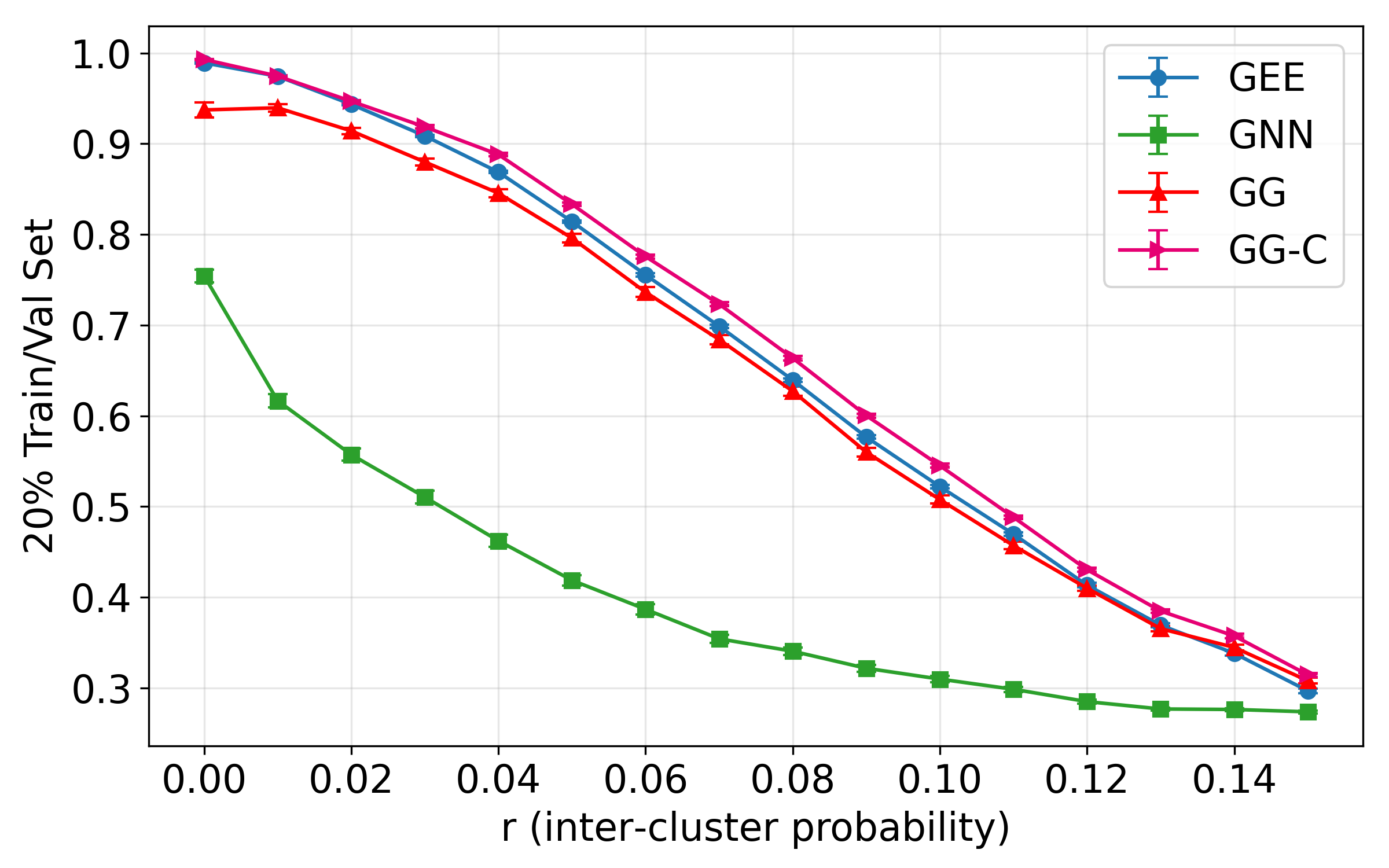}
         \label{fig:SBM_acc_20}
     }
     \hfill
     {
         \includegraphics[width=0.48\textwidth]{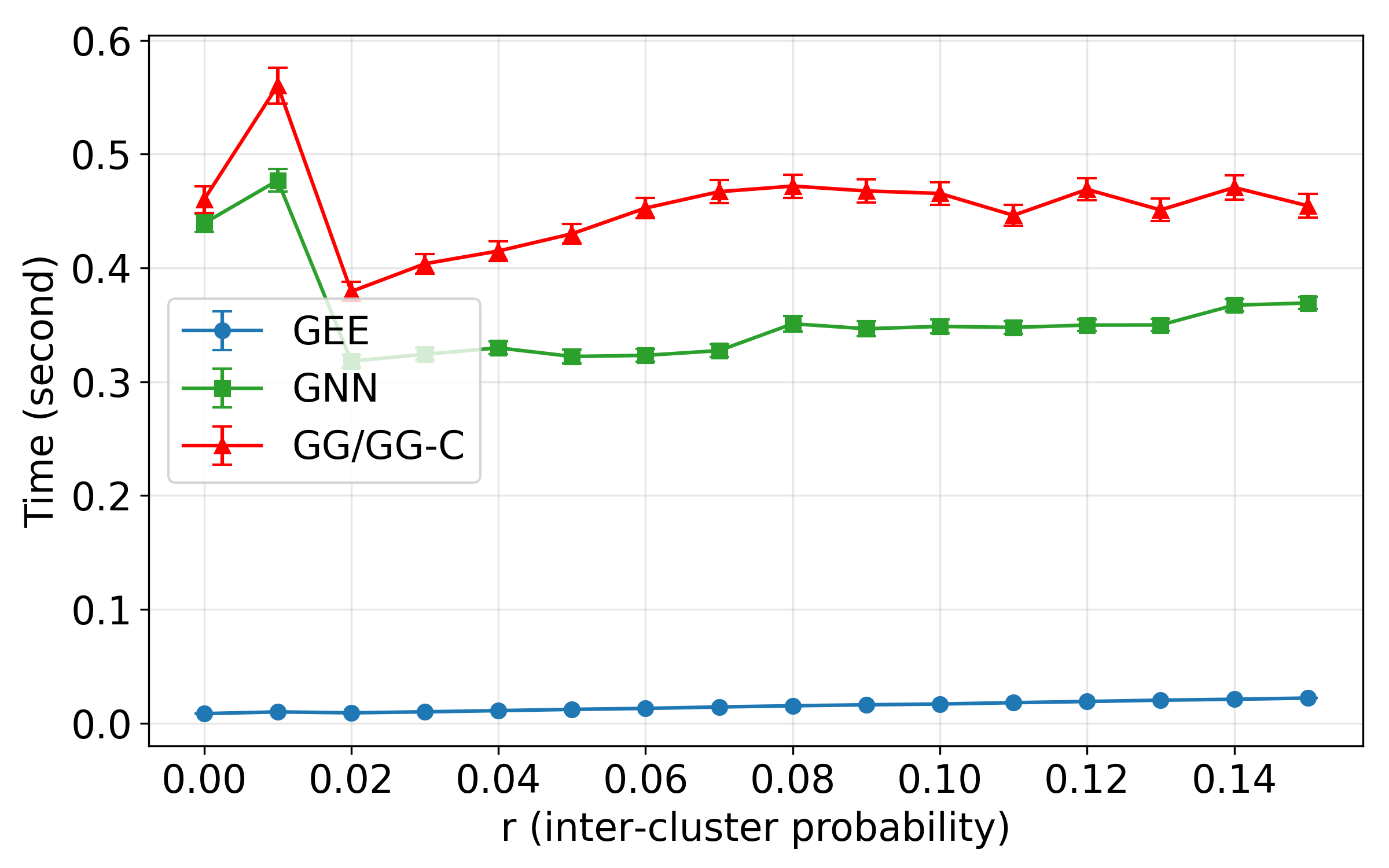}
         \label{fig:SBM_time_20}
     }
     \centering
     {
         \includegraphics[width=0.48\textwidth]{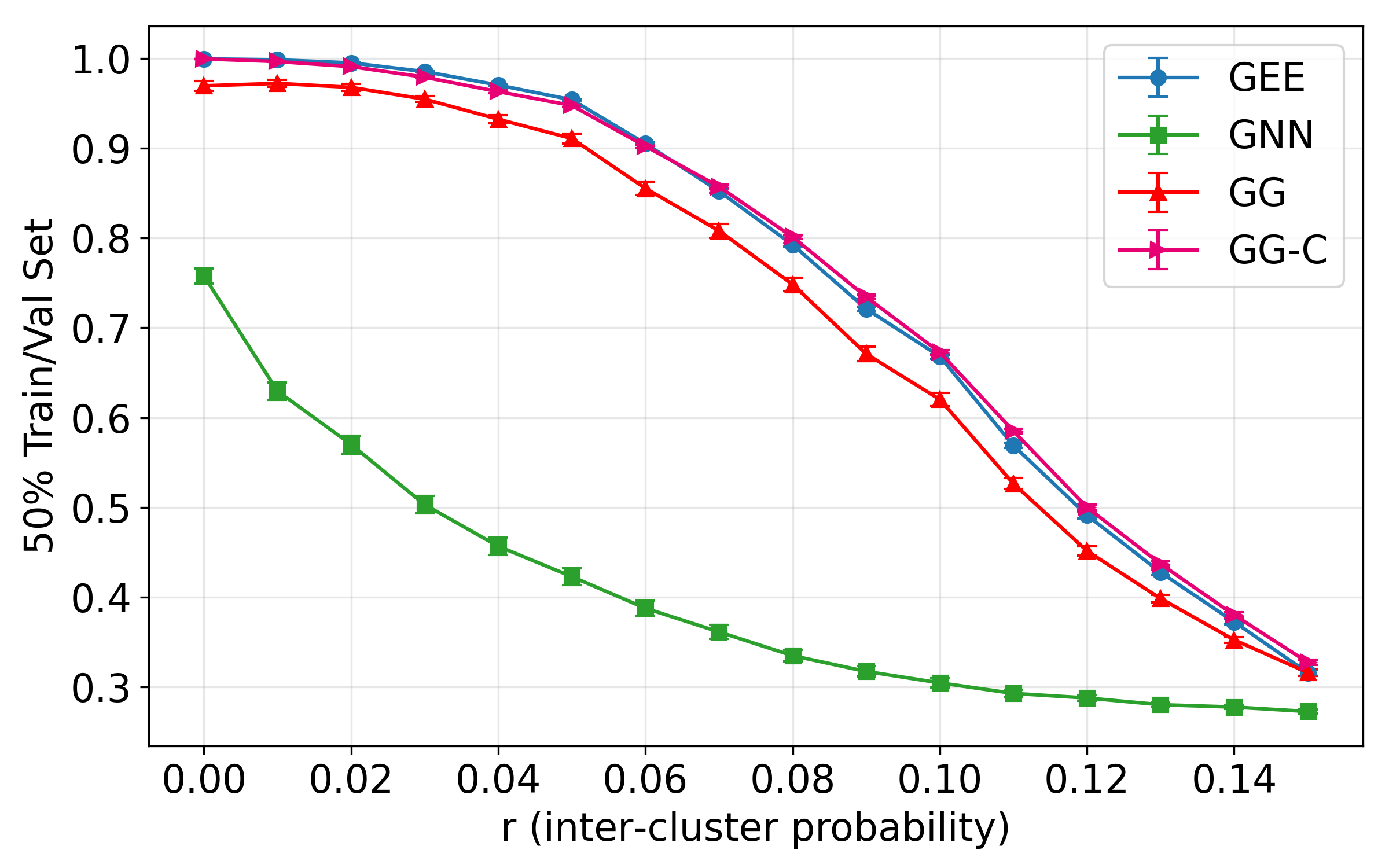}
         \label{fig:SBM_acc_50}
     }
     \hfill
     {
         \includegraphics[width=0.48\textwidth]{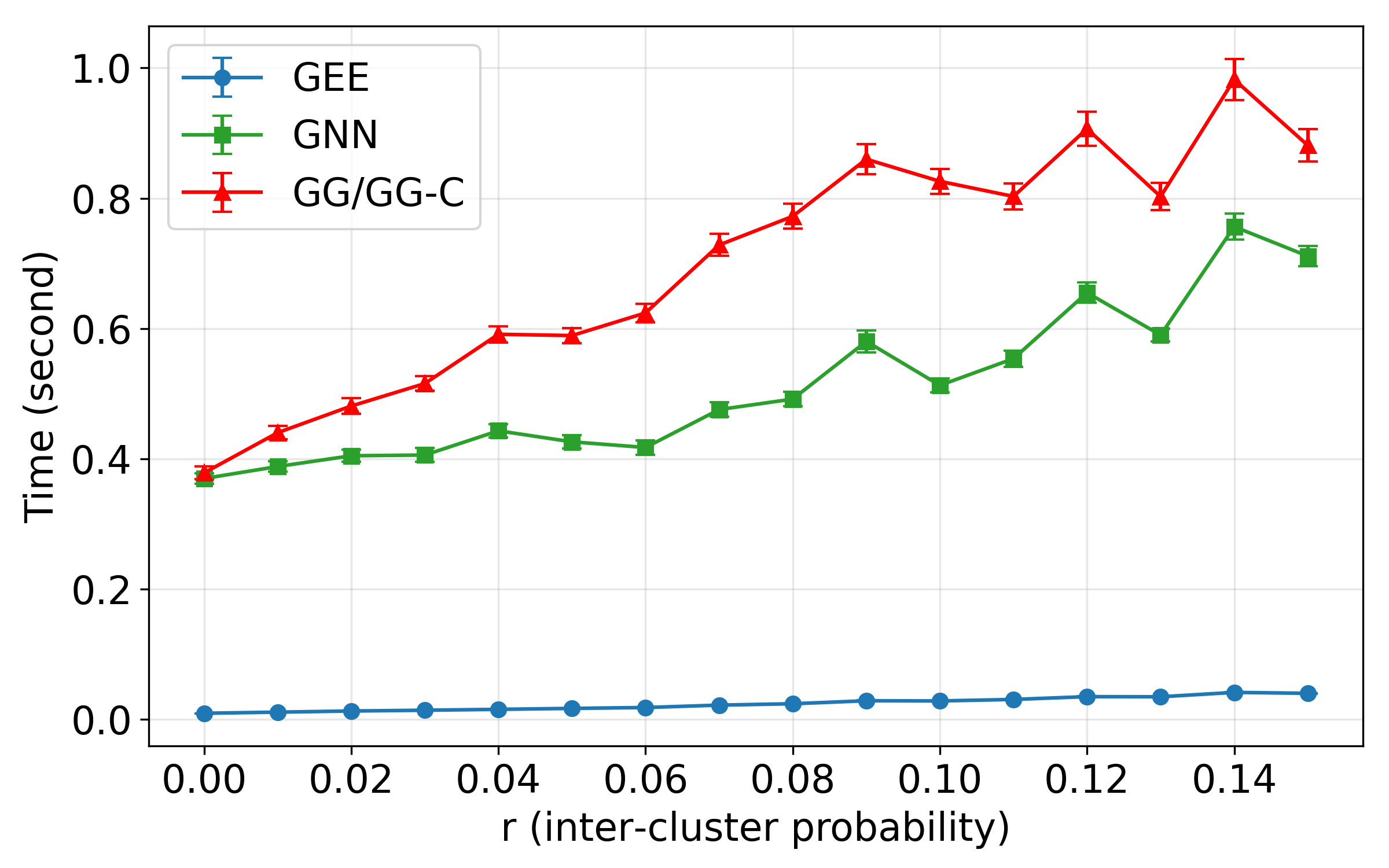}
         \label{fig:SBM_time_50}
     }
    
     \caption{
         Classification performance comparison of GEE, GNN, GG, and GG-C on SBM graphs. Each row corresponds to a different proportion of nodes used for the train/val set, ranging from 5\% (top row) to 50\% (bottom row). This figure confirms the superiority of our GG and GG-C methods. They consistently outperform the fast-but-inaccurate GNN (prone to local optima), and the advantage of GG-C over GEE is most significant in low-label settings (e.g., 5\% and 10\%), showcasing their excellent data efficiency for semi-supervised learning.
     }
     \label{fig:classification SBM all}
     \vspace{-10pt}
 \end{figure*}

 \begin{table*}[t]
   \centering
   \caption{Classification accuracy and running time on real-world datasets with 10\% (top)  and 20\% (bottom) train/val data. For accuracy, the best and second-best results are highlighted in \textbf{bold} and \textit{italic}, respectively. For the GNN-based models (GNN, GG and GG-C), the fastest running time is also marked in bold. The running time of GEE is not highlighted and is excluded from direct comparison due to its non-iterative nature.}
   \label{tab:classification realdata 10 and 20}

   \begin{subtable}[t]{\textwidth}
     \centering
     \renewcommand{\arraystretch}{1.0}
     \small
     \begin{tabular}{lcccccccc}
     \multicolumn{9}{c}{(a) Results with 10\% Train/Val Data} \\
       \toprule
       10\% & ACM & BAT & DBLP & EAT & UAT & Wiki & Gene & IIP \\
       \midrule
       \multicolumn{9}{c}{Accuracy}\\
       \midrule
       GEE  & 0.546$\pm$0.002 & \emph{0.424$\pm$0.006} & 0.353$\pm$0.001 & \textbf{0.380$\pm$0.006} & \textbf{0.473$\pm$0.003} & 0.353$\pm$0.003 & \textbf{0.618$\pm$0.003} & 0.509$\pm$0.010 \\
       GNN  & 0.464$\pm$0.005 & 0.304$\pm$0.005 & 0.309$\pm$0.003 & 0.333$\pm$0.005 & 0.377$\pm$0.004 & 0.283$\pm$0.003 & 0.544$\pm$0.006 & 0.524$\pm$0.006 \\
       GG   & \emph{0.565$\pm$0.006} & 0.392$\pm$0.008 & \emph{0.365$\pm$0.002} & 0.335$\pm$0.006 & 0.430$\pm$0.005 & \textbf{0.419$\pm$0.002} & 0.567$\pm$0.005 & \textbf{0.536$\pm$0.009} \\
       GG-C  & \textbf{0.603$\pm$0.002} & \textbf{0.427$\pm$0.007} & \textbf{0.376$\pm$0.003} & \emph{0.380$\pm$0.005} & \emph{0.467$\pm$0.003} & \emph{0.411$\pm$0.002} & \emph{0.594$\pm$0.003} & \emph{0.527$\pm$0.007} \\
       \midrule\midrule
       \multicolumn{9}{c}{Running time (s)}\\
       \midrule
       GEE      & 0.01$\pm$0.00 & 0.00$\pm$0.00 & 0.00$\pm$0.00 & 0.00$\pm$0.00 & 0.01$\pm$0.00 & 0.01$\pm$0.00 & 0.00$\pm$0.00 & 0.00$\pm$0.00 \\
       GNN      & \textbf{0.38$\pm$0.01} & \textbf{0.14$\pm$0.00} & \textbf{0.32$\pm$0.01} & \textbf{0.21$\pm$0.01} & \textbf{0.35$\pm$0.01} & \textbf{0.76$\pm$0.02} & 0.18$\pm$0.01 & \textbf{0.13$\pm$0.00} \\
       GG/GG-C   & 0.48$\pm$0.02 & 0.13$\pm$0.00 & 0.35$\pm$0.01 & 0.22$\pm$0.01 & 0.35$\pm$0.01 & 0.79$\pm$0.02 & \textbf{0.14$\pm$0.00} & 0.12$\pm$0.00 \\
       \bottomrule
     \end{tabular}
   \end{subtable}

   \vspace{1.2em}

   \begin{subtable}[t]{\textwidth}
     \centering
     \renewcommand{\arraystretch}{1.0}
     \small
     \begin{tabular}{lcccccccc}
       \toprule
       10\% & LastFM & PolBlogs & TerroristRel & KarateClub & Chameleon & Cora & Citeseer \\
       \midrule
       \multicolumn{8}{c}{Accuracy}\\
       \midrule
       GEE  & 0.444$\pm$0.002 & \emph{0.780$\pm$0.004} & \emph{0.815$\pm$0.005} & \emph{0.648$\pm$0.016} & 0.283$\pm$0.004 & 0.450$\pm$0.002 & 0.302$\pm$0.002 \\
       GNN  & 0.495$\pm$0.003 & 0.729$\pm$0.009 & 0.699$\pm$0.007 & 0.257$\pm$0.012 & \emph{0.309$\pm$0.003} & 0.359$\pm$0.005 & 0.267$\pm$0.005 \\
       GG   & \textbf{0.601$\pm$0.002} & 0.770$\pm$0.013 & 0.797$\pm$0.009 & 0.663$\pm$0.019 & 0.283$\pm$0.004 & \emph{0.478$\pm$0.005} & \emph{0.344$\pm$0.003} \\
       GG-C  & \emph{0.592$\pm$0.002} & \textbf{0.842$\pm$0.007} & \textbf{0.835$\pm$0.003} & \textbf{0.729$\pm$0.014} & \textbf{0.354$\pm$0.003} & \textbf{0.515$\pm$0.004} & \textbf{0.362$\pm$0.003} \\
       \midrule\midrule
       \multicolumn{8}{c}{Running time (s)}\\
       \midrule
       GEE      & 0.02$\pm$0.00 & 0.01$\pm$0.00 & 0.00$\pm$0.00 & 0.00$\pm$0.00 & 0.01$\pm$0.00 & 0.00$\pm$0.00 & 0.00$\pm$0.00 \\
       GNN      & \textbf{1.87$\pm$0.05} & 0.36$\pm$0.01 & \textbf{0.27$\pm$0.01} & 0.13$\pm$0.00 & \textbf{0.63$\pm$0.02} & \textbf{0.40$\pm$0.01} & \textbf{0.36$\pm$0.01} \\
       GG/GG-C   & 2.33$\pm$0.05 & \textbf{0.33$\pm$0.01} & 0.30$\pm$0.01 & \textbf{0.12$\pm$0.00} & 0.64$\pm$0.03 & 0.51$\pm$0.02 & 0.46$\pm$0.02 \\
       \bottomrule
     \end{tabular}
   \end{subtable}

   \vspace{5em}
  
   \begin{subtable}[t]{\textwidth}
     \centering
     \renewcommand{\arraystretch}{1.0}
     \small
     \begin{tabular}{lcccccccc}
       \multicolumn{9}{c}{(b) Results with 20\% Train/Val Data} \\
       \toprule
      20\%  & ACM & BAT & DBLP & EAT & UAT & Wiki & Gene & IIP \\
       \midrule
       \multicolumn{9}{c}{Accuracy} \\
       \midrule
       GEE  & 0.634$\pm$0.001 & \emph{0.410$\pm$0.008} & 0.413$\pm$0.001 & \emph{0.354$\pm$0.007} & 0.507$\pm$0.002 & 0.460$\pm$0.002 & \textbf{0.665$\pm$0.003} & 0.558$\pm$0.006 \\
       GNN  & 0.500$\pm$0.006 & 0.331$\pm$0.005 & 0.333$\pm$0.003 & 0.341$\pm$0.004 & 0.389$\pm$0.005 & 0.351$\pm$0.003 & 0.564$\pm$0.007 & 0.505$\pm$0.005 \\
       GG   & \emph{0.646$\pm$0.005} & 0.375$\pm$0.007 & \emph{0.419$\pm$0.003} & 0.299$\pm$0.005 & \emph{0.448$\pm$0.005} & \textbf{0.512$\pm$0.002} & 0.616$\pm$0.007 & \emph{0.560$\pm$0.009} \\
       GG-C  & \textbf{0.670$\pm$0.002} & \textbf{0.415$\pm$0.008} & \textbf{0.429$\pm$0.003} & \textbf{0.368$\pm$0.006} & \textbf{0.484$\pm$0.003} & \emph{0.504$\pm$0.002} & \emph{0.647$\pm$0.003} & \textbf{0.574$\pm$0.006} \\
       \midrule\midrule
       \multicolumn{9}{c}{Running time (s)} \\
       \midrule
       GEE      & 0.01$\pm$0.00 & 0.00$\pm$0.00 & 0.00$\pm$0.00 & 0.00$\pm$0.00 & 0.01$\pm$0.00 & 0.01$\pm$0.00 & 0.00$\pm$0.00 & 0.00$\pm$0.00 \\
       GNN      & \textbf{0.39$\pm$0.01} & 0.13$\pm$0.00 & \textbf{0.32$\pm$0.01} & 0.23$\pm$0.01 & \textbf{0.40$\pm$0.01} & \textbf{0.85$\pm$0.03} & 0.20$\pm$0.01 & 0.13$\pm$0.00 \\
       GG/GG-C   & 0.59$\pm$0.03 & \textbf{0.13$\pm$0.00} & 0.43$\pm$0.02 & \textbf{0.22$\pm$0.01} & 0.44$\pm$0.02 & 0.91$\pm$0.02 & \textbf{0.16$\pm$0.01} & \textbf{0.13$\pm$0.00} \\
       \bottomrule
     \end{tabular}
   \end{subtable}

   \vspace{1.2em}

   \begin{subtable}[t]{\textwidth}
     \centering
     \renewcommand{\arraystretch}{1.0}
     \small
     \begin{tabular}{lcccccccc}
       \toprule
       20\% & LastFM & PolBlogs & TerroristRel & KarateClub & Chameleon & Cora & Citeseer \\
       \midrule
       \multicolumn{8}{c}{Accuracy} \\
       \midrule
       GEE  & 0.560$\pm$0.001 & \emph{0.842$\pm$0.003} & \emph{0.865$\pm$0.002} & 0.648$\pm$0.016 & 0.279$\pm$0.004 & 0.554$\pm$0.002 & 0.385$\pm$0.002 \\
       GNN  & 0.556$\pm$0.002 & 0.746$\pm$0.012 & 0.710$\pm$0.009 & 0.257$\pm$0.012 & \emph{0.313$\pm$0.003} & 0.408$\pm$0.005 & 0.308$\pm$0.003 \\
       GG   & \textbf{0.672$\pm$0.002} & 0.816$\pm$0.012 & 0.829$\pm$0.013 & \emph{0.663$\pm$0.019} & 0.293$\pm$0.004 & \emph{0.584$\pm$0.004} & \emph{0.430$\pm$0.002} \\
       GG-C  & \emph{0.666$\pm$0.002} & \textbf{0.874$\pm$0.005} & \textbf{0.873$\pm$0.002} & \textbf{0.729$\pm$0.014} & \textbf{0.373$\pm$0.003} & \textbf{0.609$\pm$0.003} & \textbf{0.439$\pm$0.002} \\
       \midrule\midrule
       \multicolumn{8}{c}{Running time (s)} \\
       \midrule
       GEE      & 0.02$\pm$0.00 & 0.01$\pm$0.00 & 0.00$\pm$0.00 & 0.00$\pm$0.00 & 0.02$\pm$0.00 & 0.00$\pm$0.00 & 0.00$\pm$0.00 \\
       GNN      & \textbf{1.77$\pm$0.06} & 0.41$\pm$0.01 & \textbf{0.29$\pm$0.01} & 0.12$\pm$0.00 & \textbf{0.64$\pm$0.02} & \textbf{0.42$\pm$0.01} & \textbf{0.39$\pm$0.01} \\
       GG/GG-C   & 2.48$\pm$0.06 & \textbf{0.39$\pm$0.02} & 0.30$\pm$0.01 & \textbf{0.12$\pm$0.00} & 0.70$\pm$0.03 & 0.66$\pm$0.02 & 0.65$\pm$0.02 \\
       \bottomrule
     \end{tabular}
   \end{subtable}
 \end{table*}

\begin{table*}[t]
  \centering
  \caption{Full comparison of classification accuracy and running time on real-world datasets of 5\% train/val data. Suffixes (2) and (3) denote the number of GNN layers. For accuracy, the best and second-best results are highlighted in \textbf{bold} and \textit{italic}, respectively. For the GNN-based models (GNN, GG and GG-C), the fastest running time is also marked in bold. The running time of GEE is not highlighted and is excluded from direct comparison due to its non-iterative nature.}
  \label{tab:additional classification realdata}

  \begin{subtable}[t]{\textwidth}
    \centering
    \label{tab:part1A}
    \renewcommand{\arraystretch}{1.0}
    \small
    \begin{tabular}{lcccccccc}
      \toprule
      5\% & ACM & BAT & DBLP & EAT & UAT & Wiki & Gene & IIP \\
      \midrule
      \multicolumn{9}{c}{Accuracy}\\
      \midrule
      GEE & 0.469$\pm$0.002 & 0.412$\pm$0.006 & 0.316$\pm$0.001 & 0.364$\pm$0.005 & 0.415$\pm$0.005 & 0.244$\pm$0.004 & \textit{0.575$\pm$0.004} & 0.410$\pm$0.014 \\
      GNN(2) & 0.427$\pm$0.004 & 0.294$\pm$0.005 & 0.297$\pm$0.002 & 0.321$\pm$0.005 & 0.358$\pm$0.005 & 0.224$\pm$0.003 & 0.525$\pm$0.006 & 0.515$\pm$0.008 \\
      GNN(3) & 0.434$\pm$0.003 & 0.322$\pm$0.006 & 0.299$\pm$0.003 & 0.334$\pm$0.005 & 0.363$\pm$0.005 & 0.229$\pm$0.003 & 0.526$\pm$0.005 & 0.500$\pm$0.009 \\
      GG(2)  & 0.492$\pm$0.005 & 0.382$\pm$0.007 & 0.329$\pm$0.002 & 0.351$\pm$0.005 & 0.404$\pm$0.006 & \textit{0.326$\pm$0.003} & 0.531$\pm$0.005 & 0.510$\pm$0.010 \\
      GG(3) & 0.485$\pm$0.005 & 0.387$\pm$0.008 & 0.334$\pm$0.003 & 0.347$\pm$0.006 & 0.403$\pm$0.005 & \textbf{0.337$\pm$0.003} & 0.533$\pm$0.004 & \textit{0.519$\pm$0.011} \\
      GG-C(2) & \textbf{0.534$\pm$0.003} & 0.425$\pm$0.008 & 0.336$\pm$0.003 & \textit{0.370$\pm$0.005} & \textit{0.435$\pm$0.004} & 0.304$\pm$0.003 & 0.549$\pm$0.003 & 0.444$\pm$0.010 \\
      GG-C(3) & \textit{0.528$\pm$0.003} & \textit{0.430$\pm$0.007} & \textit{0.342$\pm$0.003} & \textbf{0.372$\pm$0.006} & \textbf{0.438$\pm$0.004} & 0.310$\pm$0.003 & 0.551$\pm$0.003 & 0.449$\pm$0.009 \\
      CutSSL & 0.336$\pm$0.004 & \textbf{0.447$\pm$0.005} & \textbf{0.382$\pm$0.002} & 0.365$\pm$0.002 & 0.275$\pm$0.008 & 0.042$\pm$0.002 & \textbf{0.671$\pm$0.005} & \textbf{0.636$\pm$0.002} \\

      \midrule

        \multicolumn{9}{c}{Running time (s)}\\
        \midrule
        GEE & 0.01$\pm$0.00 & 0.00$\pm$0.00 & 0.00$\pm$0.00 & 0.00$\pm$0.00 & 0.01$\pm$0.00 & 0.01$\pm$0.00 & 0.00$\pm$0.00 & 0.00$\pm$0.00 \\
        GNN($2$) & \textbf{0.37$\pm$0.01} & \textbf{0.14$\pm$0.00} & 0.30$\pm$0.01 & \textbf{0.21$\pm$0.01} & 0.32$\pm$0.01 & \textbf{0.74$\pm$0.02} & 0.16$\pm$0.00 & 0.13$\pm$0.00 \\
        GNN($3$) & 0.40$\pm$0.01 & 0.13$\pm$0.00 & 0.29$\pm$0.01 & 0.24$\pm$0.01 & 0.34$\pm$0.01 & 0.75$\pm$0.02 & 0.17$\pm$0.00 & 0.13$\pm$0.00 \\
        GG/GG-C($2$) & 0.37$\pm$0.01 & 0.14$\pm$0.00 & \textbf{0.28$\pm$0.01} & 0.22$\pm$0.01 & \textbf{0.30$\pm$0.01} & 0.77$\pm$0.02 & \textbf{0.14$\pm$0.00} & 0.13$\pm$0.00 \\
        GG/GG-C($3$) & 0.37$\pm$0.01 & 0.14$\pm$0.00 & 0.30$\pm$0.01 & 0.23$\pm$0.01 & 0.34$\pm$0.01 & 0.81$\pm$0.03 & 0.15$\pm$0.01 & \textbf{0.12$\pm$0.00} \\
        CutSSL & 11.64$\pm$0.36 & 0.33$\pm$0.01 & 5.46$\pm$0.02 & 0.77$\pm$0.02 & 8.02$\pm$0.28 & 11.31$\pm$0.40 & 0.76$\pm$0.01 & 0.38$\pm$0.02 \\

        \bottomrule
    \end{tabular}
  \end{subtable}

  \vspace{1.2em}

  \begin{subtable}[t]{\textwidth}
    \centering
    \label{tab:part2A}
    \renewcommand{\arraystretch}{1.0}
    \small
    \begin{tabular}{lcccccccc}
      \toprule
      5\% & LastFM & PolBlogs & TerroristRel & KarateClub & Chameleon & Cora & Citeseer \\
      \midrule
      \multicolumn{8}{c}{Accuracy}\\
      \midrule
      GEE & 0.333$\pm$0.004 & 0.723$\pm$0.005 & 0.737$\pm$0.010 & 0.648$\pm$0.016 & 0.267$\pm$0.004 & 0.349$\pm$0.006 & 0.244$\pm$0.002 \\
      GNN(2) & 0.431$\pm$0.003 & 0.706$\pm$0.009 & 0.683$\pm$0.006 & 0.257$\pm$0.012 & 0.288$\pm$0.003 & 0.318$\pm$0.006 & 0.238$\pm$0.003 \\
      GNN(3) & 0.444$\pm$0.003 & 0.711$\pm$0.009 & 0.681$\pm$0.008 & 0.450$\pm$0.014 & 0.295$\pm$0.003 & 0.311$\pm$0.005 & 0.245$\pm$0.003 \\
      GG(2)  & \textit{0.536$\pm$0.003} & 0.729$\pm$0.014 & 0.718$\pm$0.011 & 0.663$\pm$0.019 & 0.270$\pm$0.004 & 0.393$\pm$0.005 & 0.280$\pm$0.003 \\
      GG(3) & \textbf{0.562$\pm$0.003} & 0.744$\pm$0.015 & 0.719$\pm$0.014 & 0.697$\pm$0.015 & 0.279$\pm$0.004 & 0.404$\pm$0.006 & 0.293$\pm$0.003 \\
      GG-C(2) & 0.509$\pm$0.003 & 0.815$\pm$0.009 & \textbf{0.765$\pm$0.006} & 0.729$\pm$0.014 & \textbf{0.333$\pm$0.003} & 0.426$\pm$0.004 & 0.303$\pm$0.003 \\
      GG-C(3) & 0.531$\pm$0.003 & \textit{0.821$\pm$0.009} & \textit{0.759$\pm$0.007} & \textit{0.732$\pm$0.012} & \textit{0.332$\pm$0.003} & \textit{0.429$\pm$0.005} & \textit{0.306$\pm$0.003} \\
      CutSSL & 0.247$\pm$0.020 & \textbf{0.952$\pm$0.000} & 0.102$\pm$0.009 & \textbf{0.955$\pm$0.003} & 0.198$\pm$0.002 & \textbf{0.725$\pm$0.002} & \textbf{0.526$\pm$0.002} \\
      \midrule
        \multicolumn{8}{c}{Running time (s)}\\
        \midrule
        GEE & 0.01$\pm$0.00 & 0.01$\pm$0.00 & 0.00$\pm$0.00 & 0.00$\pm$0.00 & 0.01$\pm$0.00 & 0.00$\pm$0.00 & 0.00$\pm$0.00 \\
        GNN($2$) & \textbf{1.58$\pm$0.04} & 0.34$\pm$0.01 & 0.26$\pm$0.01 & 0.13$\pm$0.00 & 0.57$\pm$0.01 & \textbf{0.38$\pm$0.01} & \textbf{0.36$\pm$0.01} \\
        GNN($3$) & 1.66$\pm$0.05 & 0.38$\pm$0.01 & 0.32$\pm$0.03 & 0.13$\pm$0.00 & 0.64$\pm$0.02 & 0.41$\pm$0.01 & 0.38$\pm$0.01 \\
        GG/GG-C($2$)  & 2.18$\pm$0.05 & \textbf{0.30$\pm$0.01} & \textbf{0.25$\pm$0.01} & \textbf{0.13$\pm$0.00} & \textbf{0.56$\pm$0.02} & 0.39$\pm$0.02 & 0.38$\pm$0.01 \\
        GG/GG-C($3$) & 2.33$\pm$0.07 & 0.32$\pm$0.01 & 0.33$\pm$0.02 & 0.14$\pm$0.00 & 0.65$\pm$0.03 & 0.47$\pm$0.02 & 0.41$\pm$0.01 \\
        CutSSL & 53.63$\pm$1.44 & 1.60$\pm$0.03 & 4.19$\pm$0.16 & 0.16$\pm$0.00 & 16.24$\pm$0.64 & 2.95$\pm$0.04 & 5.03$\pm$0.05 \\
        \bottomrule
    \end{tabular}
  \end{subtable}
\end{table*}

} \fi

\end{document}